\documentclass[journal]{IEEEtran}

\usepackage{fancyhdr}
\usepackage[normalem]{ulem}
\usepackage[hyphens]{url}

\ifCLASSINFOpdf
\usepackage[pdftex]{graphicx}
\graphicspath{{img/}}
\DeclareGraphicsExtensions{.pdf,.png,.jpg}

\else
\usepackage[dvips]{graphicx}
\graphicspath{{img/}}
\DeclareGraphicsExtensions{.eps}

\fi

\usepackage{subfigure}
\usepackage{multirow}
\usepackage{array}
\usepackage{enumitem}
\usepackage[numbers,sort&compress]{natbib}
\usepackage{tablefootnote}
\usepackage{url}

\usepackage{mathptmx} % This is Times font
\usepackage[final]{microtype}
\usepackage{flushend}
\usepackage{siunitx}
\usepackage[usenames, dvipsnames]{color}
\usepackage[super]{nth}
\usepackage{amsmath,bm}
\usepackage[bookmarks=true,breaklinks=true,colorlinks,linkcolor=black,citecolor=blue,urlcolor=black,draft=false]{hyperref}
\makeatletter
\def\maxwidth{\ifdim\Gin@nat@width>\linewidth\linewidth\else\Gin@nat@width\fi}
\def\maxheight{\ifdim\Gin@nat@height>\textheight\textheight\else\Gin@nat@height\fi}
\makeatother
\setkeys{Gin}{width=\maxwidth,height=\maxheight,keepaspectratio}

\newcommand{\bdmath}{\begin{dmath}}
\newcommand{\edmath}{\end{dmath}}
\newcommand{\beq}{\begin{equation}}
\newcommand{\eeq}{\end{equation}}
\newcommand{\bdm}{\begin{displaymath}}
\newcommand{\edm}{\end{displaymath}}
\newcommand{\bea}{\begin{eqnarray}}
\newcommand{\eea}{\end{eqnarray}}
\newcommand{\beal}{\beq \begin{array}{ll}}
\newcommand{\eeal}{\end{array} \eeq}
\newcommand{\beas}{\begin{eqnarray*}}
\newcommand{\eeas}{\end{eqnarray*}}
\newcommand{\ba}{\begin{array}}
\newcommand{\ea}{\end{array}}
\newcommand{\bit}{\begin{itemize}}
\newcommand{\eit}{\end{itemize}}
\newcommand{\ben}{\begin{enumerate}}
\newcommand{\een}{\end{enumerate}}

\newcommand{\M}[1]{{\bm #1}} % Face for matrices
\newcommand{\hide}[1]{}

\newcommand{\hiddenText}{{\color{gray} hidden text.}}
\newcommand{\hideWithText}[1]{\hiddenText}

\newcommand{\MR}{\M{R}}

\newcommand{\vp}{\boldsymbol{p}}

\newcommand{\vv}{\boldsymbol{v}}

\renewcommand{\ij}{_{ij}}

\newcommand{\omitted}[1]{}

\newcommand{\preintRot}{\Delta \tilde{\MR}}
\newcommand{\preintVel}{\Delta \tilde{\vv}}
\newcommand{\preintTran}{\Delta \tilde{\vp}}

\begin{document}
%\title{Navion: A Fully Integrated Energy-Efficient Visual-Inertial Odometry Accelerator for Autonomous Navigation of Nano Drones}
\title{Navion: A 2mW Fully Integrated Real-Time Visual-Inertial Odometry Accelerator for Autonomous Navigation of Nano Drones}
\author{Amr~Suleiman,~\IEEEmembership{Member,~IEEE,}
        Zhengdong~Zhang,~\IEEEmembership{Student~Member,~IEEE,}
        Luca~Carlone,~\IEEEmembership{Member,~IEEE}
        Sertac~Karaman,~\IEEEmembership{Member,~IEEE}
        and~Vivienne~Sze,~\IEEEmembership{Senior Member,~IEEE}% <-this % stops a space
\thanks{A. Suleiman, Z. Zhang and V. Sze are with the Department of Electrical and Computer Science, MIT. L. Carlone and S. Karaman are with the Department of Aeronautics and Astronautics, MIT, Cambridge MA, 02139. (Project website: http://navion.mit.edu/). This work was partially funded by the AFOSR YIP FA9550-16-1-0228 and by the NSF CAREER 1350685.}}

\maketitle

% include SLAM in abstract to make more searchable
\begin{abstract}
This paper presents Navion, an energy-efficient accelerator for visual-inertial odometry (VIO) that enables autonomous navigation of miniaturized robots (e.g., nano drones), and virtual/augmented reality on portable devices. The chip uses inertial measurements and mono/stereo images to estimate the drone's trajectory and a 3D map of the environment. This estimate is obtained by running a state-of-the-art VIO algorithm based on non-linear factor graph optimization, which requires large irregularly structured memories and heterogeneous computation flow. To reduce the energy consumption and footprint, the entire VIO system is fully integrated on chip to eliminate costly off-chip processing and storage. This work uses compression and exploits both structured and unstructured sparsity to reduce on-chip memory size by 4.1$\times$. Parallelism is used under tight area constraints to increase throughput by 43\%. The chip is fabricated in 65nm CMOS, and can process 752$\times$480 stereo images from EuRoC dataset in real-time at 20 frames per second (fps) consuming only an average power of 2mW. At its peak performance, Navion can process stereo images at up to 171 fps and inertial measurements at up to 52 kHz, while consuming an average of 24mW. The chip is configurable to maximize accuracy, throughput and energy-efficiency trade-offs and to adapt to different environments. To the best of our knowledge, this is the first fully integrated VIO system in an ASIC.

%and can process 752$\times$480 stereo images at up to 171 fps and inertial measurements at up to 52 kHz, while consuming an average of 24mW at peak performance. The chip is configurable to maximize accuracy, throughput and energy-efficiency trade-offs and to adapt to different environments. By adapting to the EuRoC dataset sequences and running at the sensor's frame rate of 20 fps, Navion consumes an average power of 2mW. To the best of our knowledge, this is the first fully integrated VIO system in an ASIC.
\end{abstract}


%!TEX root=main.tex

\section{Introduction}
\begin{figure}
    \begin{center}
        \includegraphics[width=1.0\linewidth]{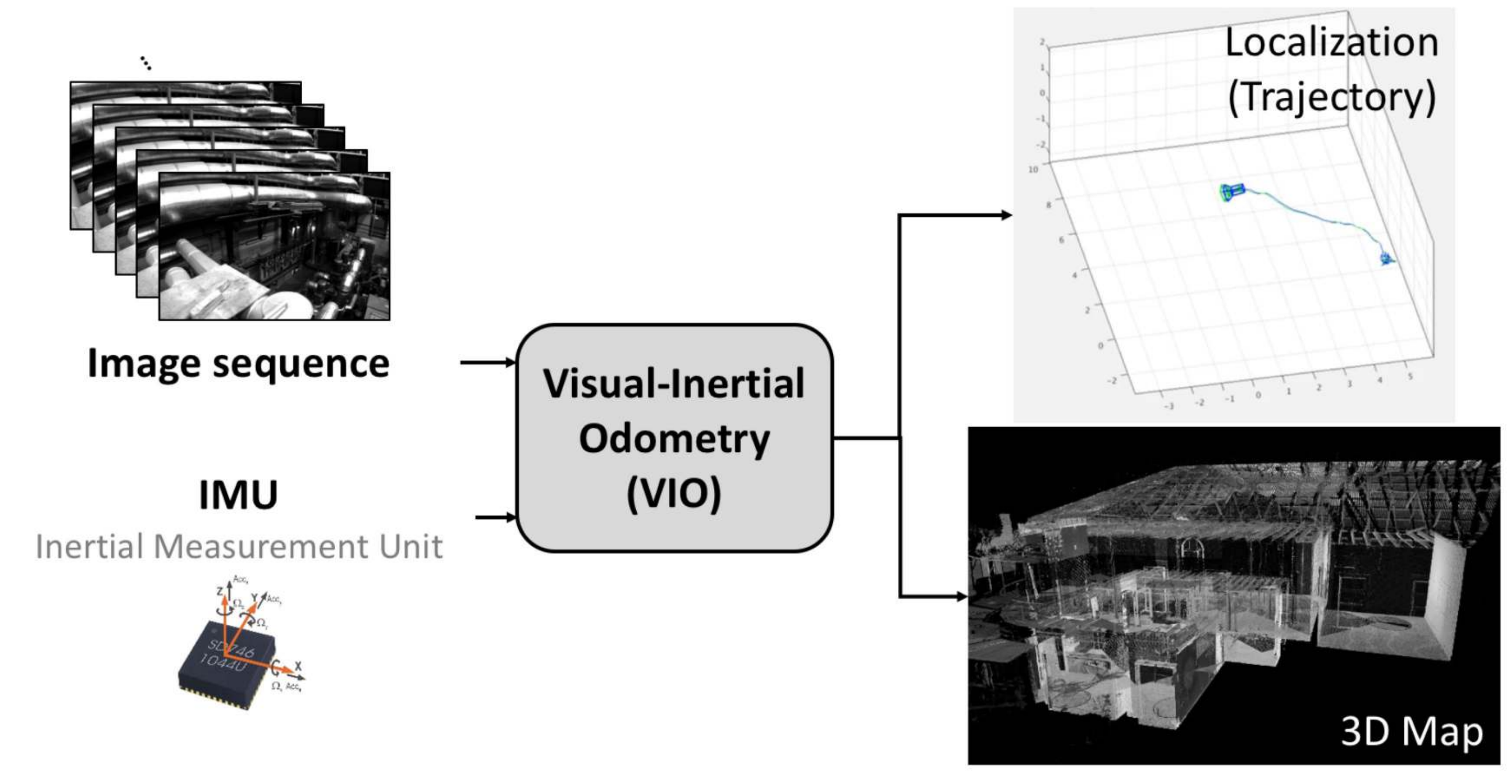}
        \caption{Visual-Inertial Odometry (VIO) processing image and IMU measurements streams for trajectory and 3D map output.}
        \label{fig:vio_basics}
    \end{center}
\end{figure}

Autonomous navigation has attracted a lot of attention with many consumer products available on the market. In this application, estimating the 3D motion within an environment, referred to as ego-motion, is of crucial importance. It is the first step towards higher level tasks including motion planning and obstacle avoidance~\cite{geomatics2014,icuas2014Mohammed,uav2015Valavanis}. Motion estimation is also important in other applications such as augmented reality (AR) and virtual reality (VR)~\cite{access2017Chatzopoulos}. This problem is well--known in the robotics community as simultaneous localization and mapping (SLAM). Visual-Inertial Odometry (VIO) is a special instance of SLAM, where visual (camera) and inertial measurement unit (IMU) data are used to estimate the location and map the surroundings. Fig.~\ref{fig:vio_basics} illustrates the inputs and outputs of a VIO pipeline. Compared with a full SLAM system, VIO does not have loop closure, which happens at much lower frequency and can be off-loaded to the cloud. As a result, a VIO system can output the estimated motion at significantly higher throughput, which is critical for autonomous navigation for fast moving robots/drones/vehicles, and also critical to reduce the motion sickness of a consumer using AR/VR devices.

Many VIO algorithms have been proposed for implementation~\cite{icra2007Mourikis,jfr2010Sibley,robotics2017Forster}. However, running these algorithms in real--time requires relatively powerful CPUs and/or GPUs~\cite{ei2015Valavanis}. Mounting powerful CPUs and GPUs on a big drone like Skydio R1 is feasible~\cite{skydio:R1}. However, such solutions cannot be applied to nano and pico drones/unmanned aerial vehicles (UAVs) as those presented in~\cite{robotics2012Wood,science2013Ma}. This is because of the constraints on the form factor as well as the extremely limited power budget available on these miniature vehicles. For example, the budget for a stable flight in nano/pico drone is around 100mW~\cite{science2013Ma}, which is an order of magnitude lower than the power dissipation of embedded CPUs~\cite{qualcommCPU}.

This motivates us to build a VIO accelerator for miniature drones. Potential utilization of custom accelerators for VIO was mentioned in a review article on pico drones, which appeared recently in Nature~\cite{nature2015Floreano}. There has been a lot of work on building energy--efficient accelerators for these applications~\cite{vlsi2013Yoon,jssc2015hong,jssc2018li}. A unified graphics and vision processor for pose estimation is presented in~\cite{vlsi2013Yoon}. However, the problem is simplified by using known markers in images which can be impractical. 
Hong et al.~\cite{jssc2015hong} implements a marker--less camera pose estimation for a practical AR application, but it depends on off--chip image processing and external storage, which increase the overall system power consumption. Li et al.~\cite{jssc2018li} implements only parts of the image processing needed in a VIO/SLAM system. One known downside of custom accelerators compared to CPU/GPU--based implementations is the lack of flexibility, which comes as a direct trade--off with the low power and fast processing that accelerators deliver. Some accelerators take it to an extreme by hard--wiring the design to perform a specific task in a specific environment~\cite{micro2016murray}, which severely limits its practical applications.

In all of these works, the hardware design is separated from the algorithmic choices; however, it is shown in~\cite{rss2017Zhang} with an FPGA prototype implementation that a co-design strategy can provide significant benefits. This work builds on~\cite{rss2017Zhang} and carries out more optimization for an efficient ASIC implementation. Here is a summary of the contributions and findings of this work:

\begin{itemize}
    \item We propose Navion, an energy-efficient and fully integrated VIO implementation on-chip that runs in real--time to enable autonomous navigation in miniaturized robots/UAVs. Compared to an optimized software VIO implementation, Navion is 1582$\times$ more energy-efficient than Xeon desktop CPU, and 684$\times$ more energy-efficient than a low power embedded ARM CPU. To the best of our knowledge, this is the first fully integrated VIO system in an ASIC.
    \item Multiple algorithmic and architectural optimizations are carried out in Navion. Efficient memory hierarchy and data movement enable a 9$\times$ reduction of the external DRAM bandwidth. Data compression reduces on-chip memory size by 4.4$\times$, while taking advantage of fixed and dynamic data sparsity enables 5.2$\times$ and 5.4$\times$  smaller memory size, respectively. Rescheduling and parallelism enable 43\% faster processing with minimal to no overhead.
    \item We find that adding sufficient adaptability in Navion's architecture can improve accuracy and increase throughput based on the environment and camera movement. Adapting Navion to the different sequences results in an additional 2.5$\times$ improvement in energy efficiency.
\end{itemize}

%We present a simple model that can identify the difficulty of an image sequence to adapt Navion accordingly. Moreover, adapting 

The rest of the paper is organized as follows: Section~\ref{sec:algorithms} gives an overview of the VIO algorithm. In Section~\ref{sec:architecture}, Navion architecture is presented showing the degree of flexibility in the architecture. Section~\ref{sec:optimizations} discusses some of the hardware optimizations for memory size reduction. Section~\ref{sec:results} shows the evaluation of Navion's accuracy and an analysis of the effect of changing the chip's configuration parameters on overall performance. Finally, Section~\ref{conclusion} concludes the paper.

%!TEX root=main.tex

\section{Overview of Visual-Inertial Odometry (VIO)}
\label{sec:algorithms}
\begin{figure*}
    \begin{center}
        \includegraphics[width=1.0\linewidth]{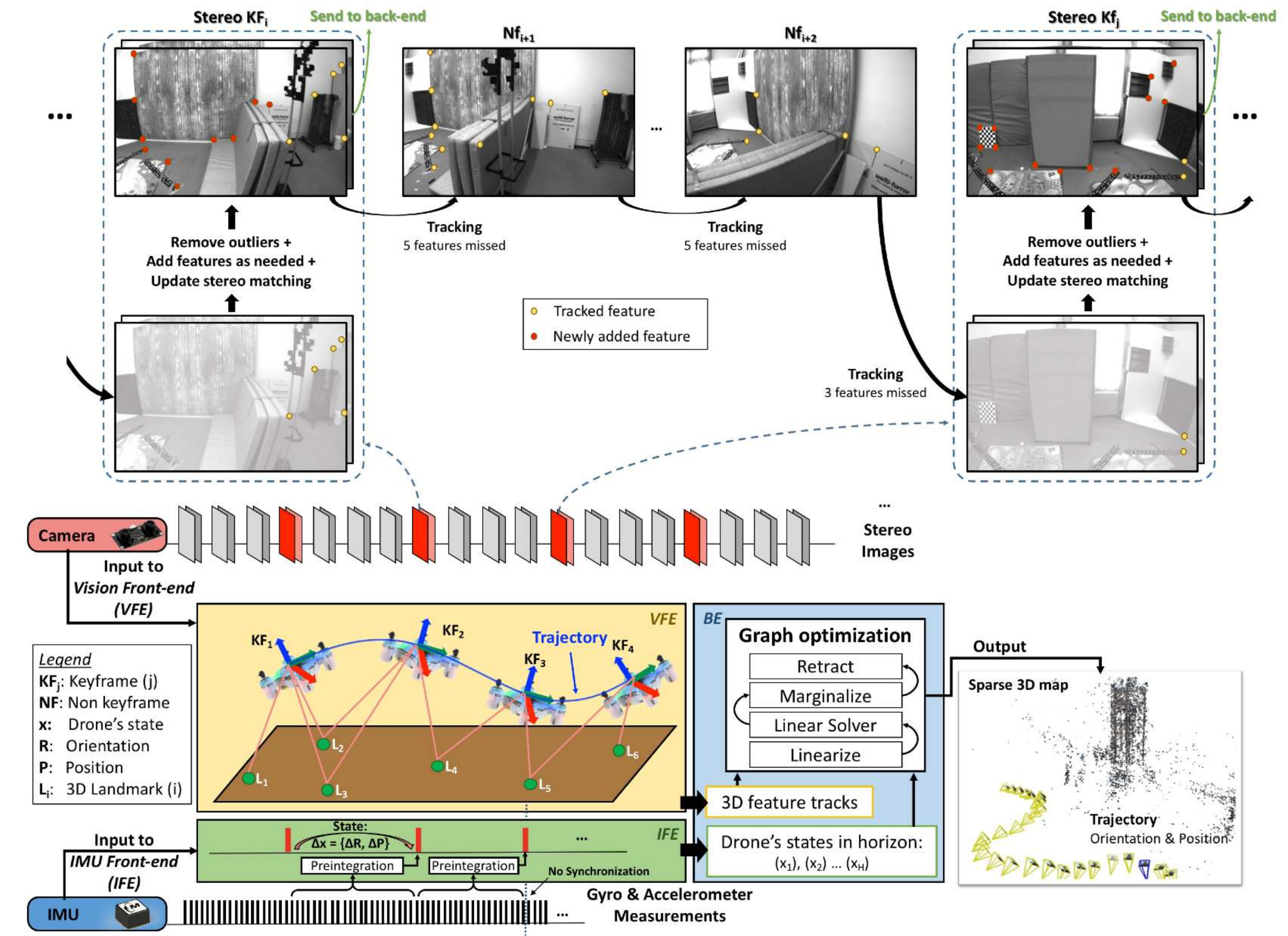}
        \caption{VIO pipeline based on~\cite{rss2015Forster}. $IFE$ summarizes the IMU measurements between $KF$s, while $VFE$ tracks features in input images that correspond to important landmarks in the scene. $VFE$ is $KF$ based, i.e., it tracks features in all frames, but removes outliers, adds new features to compensate lost ones, and performs stereo matching only at $KF$s. $BE$ fuses $VFE$ and $IFE$ outputs to perform state estimation at $KF$s using factor graph optimization.}            
        \label{fig:vio_pipeline}
    \end{center}
\end{figure*}

The Visual-Inertial Odometry (VIO) can be used to estimate the trajectory of a sensing device (e.g., sensors mounted on a drone) while reconstructing a map of the environment as shown in Fig.~\ref{fig:vio_pipeline}. The trajectory is the collection of the drone's state $x$, specifically its position $P$ and orientation $R$, over time. These states are estimated based on measurements of the environment. Navion implements the keyframe-based VIO pipeline described in~\cite{robotics2017Forster}. The keyframes ($KF$s) are a subset of the incoming frames at which the state estimation is performed. $IMU$ is used for a continuous motion/state estimation in between $KF$s. The pipeline consists of three main components: Vision frontend ($VFE$), IMU frontend ($IFE$), and Backend ($BE$) following the standard terminology in~\cite{robotics2016Cadena}. The $VFE$ tracks 3D landmarks ($L_i$) in the scene by extracting their corresponding 2D features (i.e., corners) from a camera frame, and tracks them between consecutive frames to create \textit{feature tracks}. The $IFE$ summarizes the IMU sensor data between two camera $KF$s into one measurement. The $BE$ fuses the summarized inputs from the two sensors through a non-linear graph optimization, and then it outputs the position and orientation of the drone. 

\subsection{Vision Frontend (VFE)}
The $VFE$ detects and tracks 2D features that correspond to the 3D landmarks in the scene across frames from the camera. It supports both mono and stereo modes. $VFE$ implements five main functions on the input images: feature tracking ($FT$), feature detection ($FD$), undistort \& rectify ($UR$), stereo matching ($SM$), and geometric verification ($GV$). 

\begin{itemize}
\item \textbf{Feature Tracking ($FT$)} uses the Pyramidal Lucas-Kanade optical flow~\cite{ai1981Lucas} to track features corresponding to the same landmark across all frames, including non $KF$s.

\item \textbf{Feature Detection ($FD$)} extracts features in the left stereo frame using the lightweight Shi-Tomasi corners~\cite{cvpr1994Shi}. Features are detected on a grid to keep the number of features in different regions of a $KF$ constant.

\item \textbf{Undistort \& Rectify ($UR$)} processes the left and right stereo frames only at $KF$s to prepare them for disparity calculations.

\item \textbf{Stereo Matching ($SM$)} is only active in stereo mode and it only runs on $KF$s. It gets the 3D coordinate of a feature using template matching between left and right stereo frames~\cite{Hartley2004}. Since stereo frames are undistorted and rectified, template matching happens on a horizontal search line.

\item \textbf{Geometric Verification ($GV$)} removes any incorrect matches (i.e., outliers) resulting from $FT$ and/or $SM$, and it runs only at $KF$s. Random sample consensus ($RANSAC$) is used in this pipeline. In Navion, $RANSAC$ is informed by the rotation estimate produced by gyroscope integration. This enables using the simpler 2-point method~\cite{bmvc2011Kneip} for $FT$ (i.e., mono $RANSAC$) and 1-point method~\cite{robots2009Civera} for $SM$ (i.e., stereo $RANSAC$), instead of the more complicated 5-point $RANSAC$ method.
\end{itemize}

\textit{In summary}, $VFE$ processes the input mono/stereo frames to track several landmarks ($L_i$ in Fig.~\ref{fig:vio_pipeline}) based on their 2D projections in the images (i.e., features $f_i$ in Fig.~\ref{fig:vio_pipeline}). $VFE$ outputs landmark IDs, and their corresponding feature coordinates in left and right frames; this information will be used to generate feature tracks in the $BE$. In Navion, $VFE$ uses feature tracking rather than feature descriptors to find features correspondence between frames. This avoids the computational overhead of the costly feature matching process.

%\begin{figure}
%    \begin{center}
%        \includegraphics[width=1.0\linewidth]{figures/Fig_vision_frontend_process.pdf}
%        \caption{Vision frontend process. Images from EuRoC dataset~\cite{roboticss2016Burri}.}
%        %\vspace{-10pt}                
%        \label{fig:vision_frontend_process}
%    \end{center}
%\end{figure}

\subsection{IMU Frontend (IFE)}
The $IFE$ summarizes all the IMU measurements between two consecutive $KF$s into a single preintegrated measurement.  The mathematical details are outside the scope of this paper and can be found in~\cite{rss2015Forster}. The preintegration between $KF$s \textit{i} and \textit{j} gives the relative rotation ($\preintRot\ij$), velocity ($\preintVel\ij$) and position ($\preintTran\ij$). At each $KF$, $IFE$ outputs the preintegration measurements to $BE$.

\subsection{Graph Optimization Backend (BE)}
\label{sec:factor_graph_algo}

The $BE$ fuses the feature tracks from the $VFE$ output and the IMU data from the $IFE$ by solving a non-linear optimization problem whose solution is the $KF$ state estimation, which describes the drone's trajectory and a sparse 3D landmarks map. The optimization is performed on multiple states within a sliding time window, referred to as the \emph{horizon}, using a fixed-lag smoother to compute the maximum a posterior ($MAP$) estimate~\cite{rss2015Forster}. It formulates the problem as a \textit{factor graph}~\cite{it2001Kschischang}, which describes the relationship and thus constraints\footnote{Here, we don't use the formal definition of constraints for optimization. Instead, these factors are included in the objective function in the optimization.} between all states in the horizon; these relationships between states defined by the factors are non-linear.  There are three main types of factors in $BE$: \emph{vision factors} are the feature tracks that describes the constraints between the 3D position and orientation of all the $KF$s that observe the same landmark; \emph{IMU factors} describes the difference between the relative motion between a pair of $KF$s given by the IMU measurements and the estimated relative motion from the state variables, and \emph{marginalization} factor the describes the constraints between the states of the $KF$s that were connected by any factor that fall outside the current horizon.  The $BE$ must perform an optimization to find the $KF$ states that best satisfies the constraints set by these factors and minimize any discrepancies. 

$BE$ solves the minimization problem using an on manifold Gauss-Newton method~\cite{rss2015Forster}. The four main functions are \textit{linearization}, \textit{linear solver}, \textit{marginalization}, and \textit{retract} as shown on the right side of Fig.~\ref{fig:vio_pipeline}. Specifically, the factors are linearized and accumulated into a large system of linear equations (\textit{H$\Delta$x=$\epsilon$}), where the Hessian matrix $H$ and the error vector $\epsilon$ describe how the frontend measurements affect each $KF$ state in the horizon, and the solution vector $\Delta$x is the state update. 

A linear solver uses Cholesky factorization and back-substitution to solve this linear system of equations~\cite{robotics2008Kaess}. Marginalization summarizes the information of the states that fall outside of the current time window. All the factors that constrain these states are removed from the graph.  As a result of marginalization and limited horizon, the structure of the factor graph is dynamically updated over time. Finally, the solution of the linear system ($\Delta$x) is used to update the remaining $KF$ states in the horizon using retract such that the updated variable is also on the manifold. Generally, the Gauss-Newton method is iterative but our BE runs only one iteration because the IMU information is used to provide a good initialization.

Navion's $BE$ supports a very large and continuously varying factor graph problem, with a maximum of 20 $KF$s in the horizon and 4000 vision factors (i.e., landmarks); the selection of these parameters are described next in Section~\ref{sec:parameters}. This poses challenges in storing and maintaining the graph on--chip efficiently.

\subsection{Design Considerations}
\label{sec:parameters}
\paragraph{Parameters} In Navion, a few important parameters can have significant impact on the trade--off in memory, throughput and accuracy of the design.

\begin{itemize}
    \item \textbf{Horizon size} determines how many $KF$s are used in the $BE$. Let $N$ be the number of $KF$s in a horizon. The size of the linear system that $BE$ solves after linearization is $O(N^2)$. The time complexity for solving this linear system is $O(N^3)$. Therefore, reducing $N$ can significantly reduce the memory and increase the throughput of the system, but it will hurt the accuracy of the estimated trajectory~\cite{rss2017Zhang}. Navion has a horizon size of 20.
    \item \textbf{Number of features tracked per frame} affects all the major components of $VFE$ linearly in terms of the power and the throughput. Navion tracks up to 200 features per frame.
    \item \textbf{Feature track age} is the maximal length allowed for a feature track. This parameter affects how many feature tracks will be included in the factor graph based optimization. Since each feature corresponds to a vision factor, scaling the feature track age affects the $BE$ linearly. Note that a larger feature track age is generally preferred for high accuracy. Navion has a feature track age of 10.
\end{itemize}

In applications where the tolerance of localization error is higher, it is possible to tune these parameters to reduce the memory and increase the throughput of the design significantly.

\paragraph{Optimization Method}
We employ a non--linear factor-graph based optimizer rather than an extended Kalman Filter ($EKF$) in our $BE$. $EKF$ essentially summarizes all the feature tracks between two adjacent $KF$s into one factor, hence the maximal feature age can be considered to be 2. Therefore, $EKF$ has lower complexity but its average motion (i.e., state) estimation error can be $3\times$ higher when benchmarked on the EuRoC dataset~\cite{delmerico2018benchmark}. 

Bundle adjustment is also a very well known technique to estimate the camera positions from the visual features~\cite{Hartley2004}. The goal is to minimize the total reprojection error from the landmark to the features. It is identical to a special case of the factor graph optimizer presented in this work, where only vision factors and prior factors are used. 

\paragraph{VIO vs VO}
Including inertial measurements increases the robustness of a SLAM system, especially in the indoor environments where insufficient visual features can be detected and tracked (e.g., a drone facing a white wall). It can also provide a good initialization for the non--linear optimization in the backend, reducing the number of iterations needed for $BE$ to converge to a good state estimation.

\paragraph{Stereo vs Mono}
Using stereo frames avoids the scale ambiguity of a mono VIO system~\cite{Hartley2004}. However, a stereo camera has higher power consumption than a mono camera. Therefore, Navion supports both mono and stereo modes for the user to choose depending on the target application scenario.

%!TEX root=main.tex

\section{Navion: Architecture}
\label{sec:architecture}

\begin{figure*}
    \begin{center}
        \includegraphics[width=1.0\linewidth]{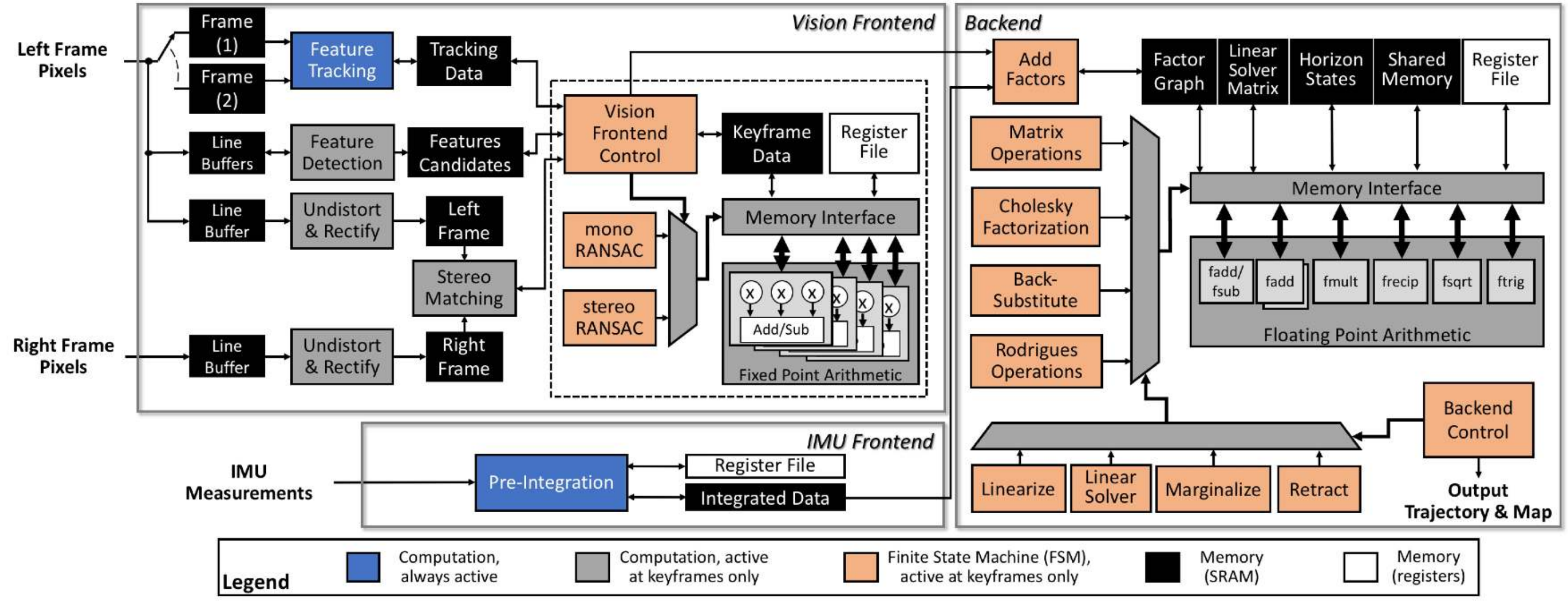}
        \caption{Overall architecture of Navion chip. All components of the VIO pipeline are integrated on--chip. Frame buffers, Factor Graph, and Linear Solver memories account for 95\% of the on--chip memory. Memory optimizations, along with rescheduling and parallelism, are carried out to enable full integration.}
        \label{fig:navion}
    \end{center}
\end{figure*}

Fig.~\ref{fig:navion} shows the overall architecture of the proposed VIO hardware accelerator. The VIO pipeline is fully integrated in Navion with no external storage or computation required. The architecture consists of the frontend (i.e., $VFE$, $IFE$), which processes the high--dimensional sensor data, and the backend (i.e., $BE$) which performs the state estimation following the algorithm discussed in Section~\ref{sec:algorithms}. Navion is a standalone accelerator that takes images (mono or stereo) and inertial measurements as inputs, and outputs the trajectory and the sparse 3D map of the surrounding environment.

With the optimizations carried out in this work, no external storage is required (e.g., DRAM), which lowers the overall system power consumption. Navion has more than 250 programmable parameters (a subset of them is shown later in Table~\ref{tab:max_params}). Some of these parameters are used to define the camera and IMU calibration and settings, such that a wide range of sensors can be used with Navion. The remaining parameters are used to control the VIO pipeline's complexity and trade--off accuracy, throughput and power consumption across different environments as shown later in Section~\ref{sec:adapt}.

\subsection{IMU Frontend (IFE)}
Inertial measurements, with 6 values per measurement (i.e., 3 values from the accelerometer and 3 values the gyroscope) are streamed into the chip through a 32--bit input bus in single precision. $IFE$ (lower left box in Fig.~\ref{fig:navion}) processes them using double precision arithmetic to perform the IMU preintegration based on~\cite{rss2015Forster}. It contains a small shared register file and a 2kB SRAM to store intermediate data and the preintegrated results. $IFE$ uses a dedicated double precision arithmetic compute unit that runs in parallel with the rest of the chip. The number of operations in the preintegration process is relatively small, hence $IFE$'s complexity is low compared to the rest of the chip. It accounts for only 2.4\% of the chip area, while consuming around 1\% of the average total power.

\subsection{Vision Frontend (VFE)}
\label{sec:vfe_arch}
The input images, with a maximum resolution of 752$\times$480 pixels, are streamed into the chip through two 8--bit buses for left and right stereo images, which are processed by the $VFE$ (upper left box in Fig.~\ref{fig:navion}). The input pixels are stored on--chip in line and/or frame buffers to lower the external bandwidth by up to 9$\times$ to 0.34 MB/frame only, which is the minimum bandwidth required to stream in a single frame (i.e., no pixel is read more than once). $VFE$ uses fixed point arithmetic for lower power and smaller area, and can be divided into two different dataflows: parallel image processing and serial geometric validation.

The image processing dataflow (i.e., $FT$, $FD$, $UR$, and $SM$) processes pixels in parallel to increase the throughput. The geometric validation, highlighted by the dashed box inside $VFE$ in Fig.~\ref{fig:navion}, implements mono and stereo $RANSAC$ using finite state machines ($FSM$), shared memory and shared arithmetic compute unit. $VFE$ accounts for 53\% of the overall chip area, and contains 439.4kB SRAM which is dominated by four frame buffers: original frames from the camera are stored in \textit{Frame (1)} and \textit{Frame (2)} memories to support $FT$. Undistorted and rectified frames are stored in \textit{Left Frame} and \textit{Right Frame} memories to support $SM$. $VFE$ detects a total of 1824 features per frame, and can track up to 200 features per frame.

\subsection{Backend}
$BE$ (right box in Fig.~\ref{fig:navion}) solves the factor graph optimization problem. Due to the serial nature of this process, a dataflow similar to the geometric validation in $VFE$ is used with a complex $FSM$. To reduce area and increase resource sharing, $BE$'s $FSM$ is divided into a hierarchy of smaller $FSM$s. The small $FMS$s include: 1) Matrix operations that are used all over the factor graph optimization problem. 2) Cholesky factorization and back--substitution used in both linear solver and marginalization processes. 3) Rodrigues operations that are used in the linearization and retract processes.

A shared register file is used, with 85 double precision registers, as a memory hierarchy for intermediate data storage similar to $IFE$, along with 412.6kB of SRAM. The \textit{Factor Graph} memory stores the $VFE$ and $IFE$ outputs in the horizon, which are then linearized into the \textit{Linear Solver Matrix} memory. $BE$ includes more than 4000 factors to support a maximum horizon size of 20 $KF$s in the optimization problem. \textit{Horizon States} memory stores the state of the $KF$s in the horizon, while the \textit{Shared Memory} stores large intermediate data. $BE$ outputs the trajectory and the sparse 3D map of the surroundings at the $KF$ rate.

Navion uses double precision arithmetic in the $BE$ and $IFE$ to ensure robustness of the VIO system. This doubles the size of all memories in $BE$ and $IFE$. Using single precision does not provide sufficient numerical precision for the underlying VIO algorithm from~\cite{rss2015Forster}. This is due to the fact that the VIO pipeline is an open--loop system. Therefore, as time goes on, the uncertainty of the state estimation keeps increasing and the condition number of the linear system increases; this causes the problem to be ill--defined and thus cannot be properly solved.
% So, some optimizations are carried out to reduce this size as shown later in Sections~\ref{sec:unstructured_sparsity} \&~\ref{sec:structured_sparsity}.
%We observed that the uncertainty of with single precision increases significantly faster than double precision. 

\subsection{Processing Modes}
Navion has two modes of operations since the VIO is $KF$--based, as shown earlier in Fig.~\ref{fig:vio_pipeline}. A detailed timing diagram is shown later in Fig.~\ref{fig:timing_breakdown}. $IFE$ is active in both modes.

\textbf{KF processing} (stereo images labeled red in Fig.~\ref{fig:vio_pipeline}): In this mode, stereo frames are streamed in and all components are active. $FT$, $FD$, $UR$ and $SM$ modules process the incoming frames in parallel. The control $FSM$ then starts the mono and stereo $RANSAC$ in series to perform the geometric verification and remove outliers before adding new features as needed. $BE$ then starts solving the factor graph optimization problem. The trajectory and sparse map outputs are updated after $BE$ is done.

\textbf{Non--KF processing} (all other stereo images in Fig.~\ref{fig:vio_pipeline}): In this mode, $FT$ is the only active component, while the rest of $VFE$ and $BE$ are off and clock--gated to reduce their power consumption. Right frames are not streamed in to reduce the off--chip memory bandwidth. The previously tracked features are stored in the \textit{Tracking Data} memory, and $FT$ updates them in--place if successfully tracked, or removes them from the memory if tracking fails. On average, processing non--$FK$ is 3.8$\times$ faster than processing $KF$.

%\begin{figure}
%    \begin{center}
%        \includegraphics[width=1.0\linewidth]{figures/Fig_KF_timing.pdf}
%        \caption{Timing diagram of keyframe processing.}            
%        \label{fig:KF_timing}
%    \end{center}
%\end{figure}

%The remainder of this section shows the various optimization on both algorithmic and architectural levels that are carried out for an energy efficient ASIC implementation.

\section{Architecture Optimizations}
\label{sec:optimizations}

This section presents the main optimizations carried out to enable energy--efficient full VIO integration in Navion. Table~\ref{tab:memories}, shown in next section, summarizes the resulting size reduction of the various on--chip memories, which is important to reduce power consumption and area cost. Additionally, we exploit parallelism and rescheduling to increase the overall throughput.

\subsection{Image Compression}
$VFE$ has four frame buffers as shown in Section~\ref{sec:vfe_arch}. These buffers are needed because of the random reading patterns in $FT$, as a feature can move to anywhere between frames depending on the camera movement. Additionally, multiple frame readings happens over time in $SM$ since it runs twice at different time slots. As a result, frames are stored on--chip to avoid re--streaming. Lossy image compression is used to reduce the size of these frame buffers. 

\begin{figure}
    \begin{center}
        \includegraphics[width=0.8\linewidth]{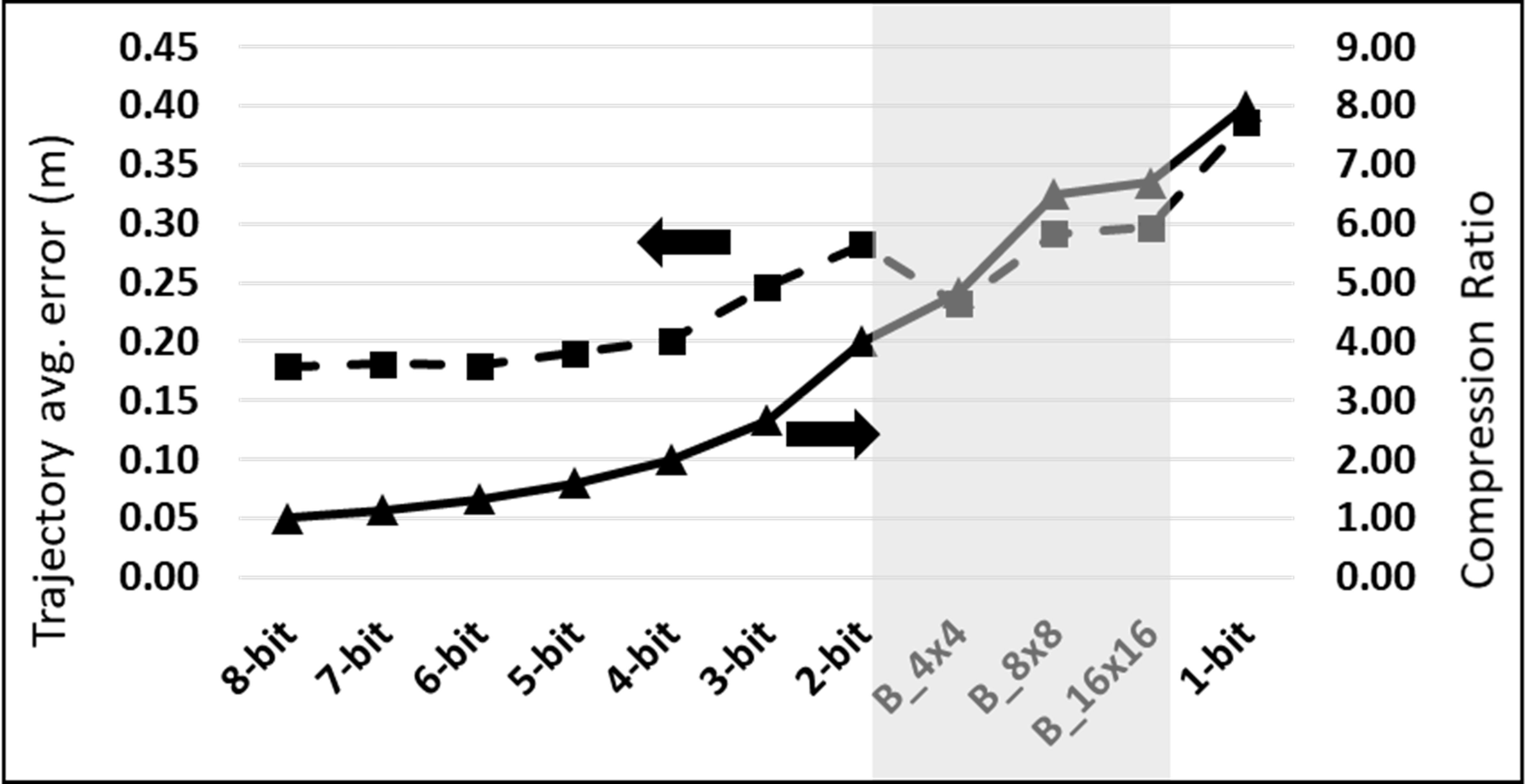}
        \caption{Memory savings vs. VIO error with lossy image compression.}
        \label{fig:compression_sweep}
    \end{center}
\end{figure}

Fig.~\ref{fig:compression_sweep} shows the trade--off between the compression ratio and the VIO error. For a minimal overhead cost, two compression methods are analyzed. Truncating the least significant bits ($LSB$) the simplest method to reduce the bitwidth with no overhead. Fig.~\ref{fig:compression_sweep} shows that VIO error increases by just 6\% going from 8--bit to 5--bit per pixel while achieving 38\% memory size reduction. However, the error increases much faster after that to more than double at the extreme of 1--bit per pixel.
    
Another lossy compression technique is block--wise quantization. The frame is divided into blocks of $N$$\times$$N$ pixels, and the dynamic range of each block is quantized into 2 levels for a 1--bit per pixel representation. This technique's overhead includes a line buffer storing $N-1$ rows and some logic to calculate the dynamic range of each block. The shaded part in Fig.~\ref{fig:compression_sweep} shows the numbers for three different block sizes: 4$\times$4, 8$\times$8, and 16$\times$16. The 4$\times$4 block quantization achieved less error than the 3--bit case, while increasing the memory savings from 2.7$\times$ to 4.4$\times$.

\begin{figure}
    \begin{center}
        \includegraphics[width=1\linewidth]{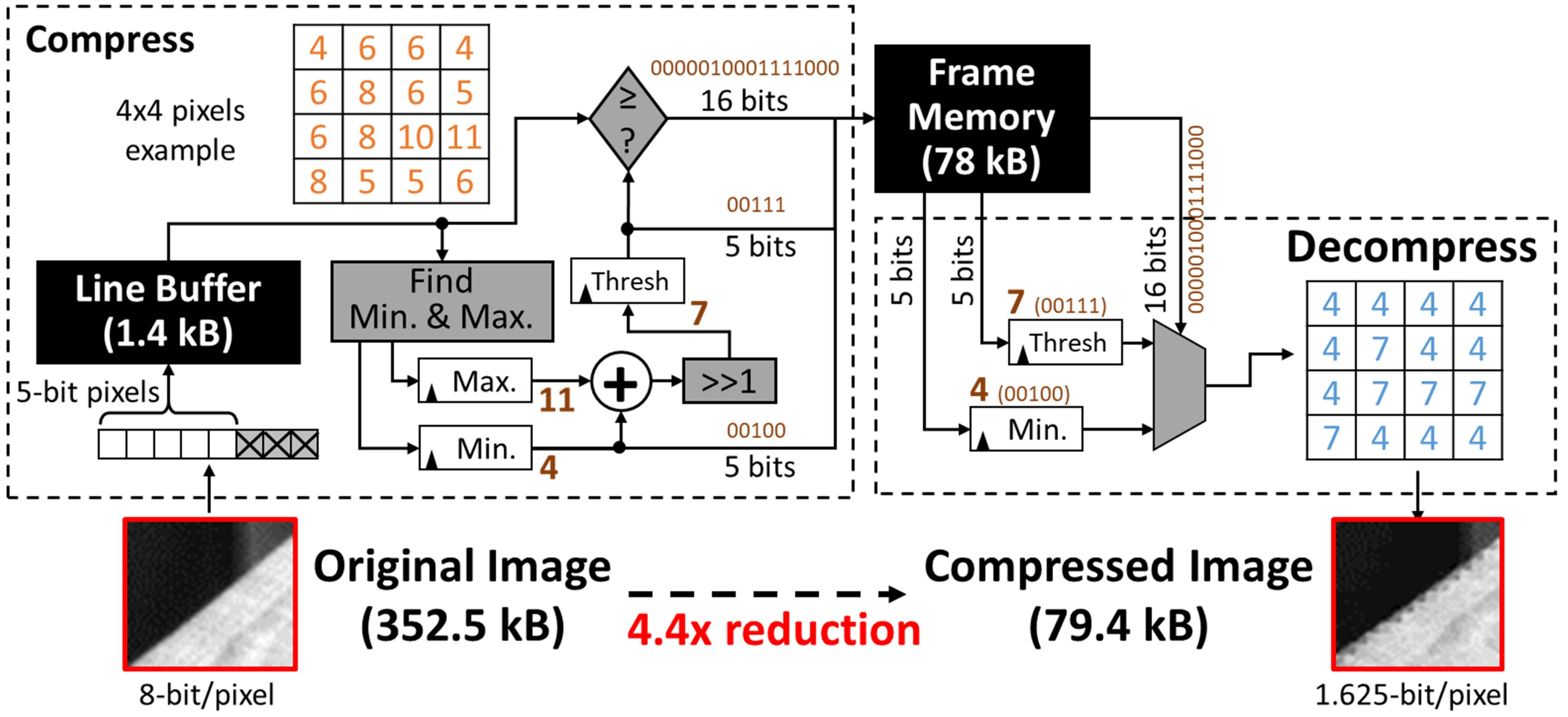}
        \caption{Block--wise image compression architecture.}
        \label{fig:image_compression}
    \end{center}
\end{figure}

Fig.~\ref{fig:image_compression} shows the image compression architecture used in Navion, which combines both lossy compression techniques described above. The pixels are quantized to 5--bit by simple $LSB$ truncation, then the image is divided into blocks of 4$\times$4 pixels. The pixel intensity dynamic range within each block is found, and a threshold divides this range in half. Every pixel is then represented by 1--bit, which is the result of comparing the pixel intensity to the threshold. Accordingly, each 4$\times$4 block of pixels uses 26 bits to store its dynamic range (1-bit/pixel), threshold (5-bit) and minimum value (5-bit). With a compute overhead of 4 kgates (0.8\% of the total kgates) and a 1.4kB line buffer, compression reduces the frame memory size by 4.4$\times$ and power by 4.9$\times$. Compressed frames are used in $SM$ and $FT$, but they are not use in $FD$ because it is more sensitive blocking and quantization artifacts as shown in Fig.~\ref{fig:image_compression}.

\subsection{Feature Tracks Unstructured Sparsity}
\label{sec:unstructured_sparsity}
\begin{figure}
    \begin{center}
        \includegraphics[width=1.0\linewidth]{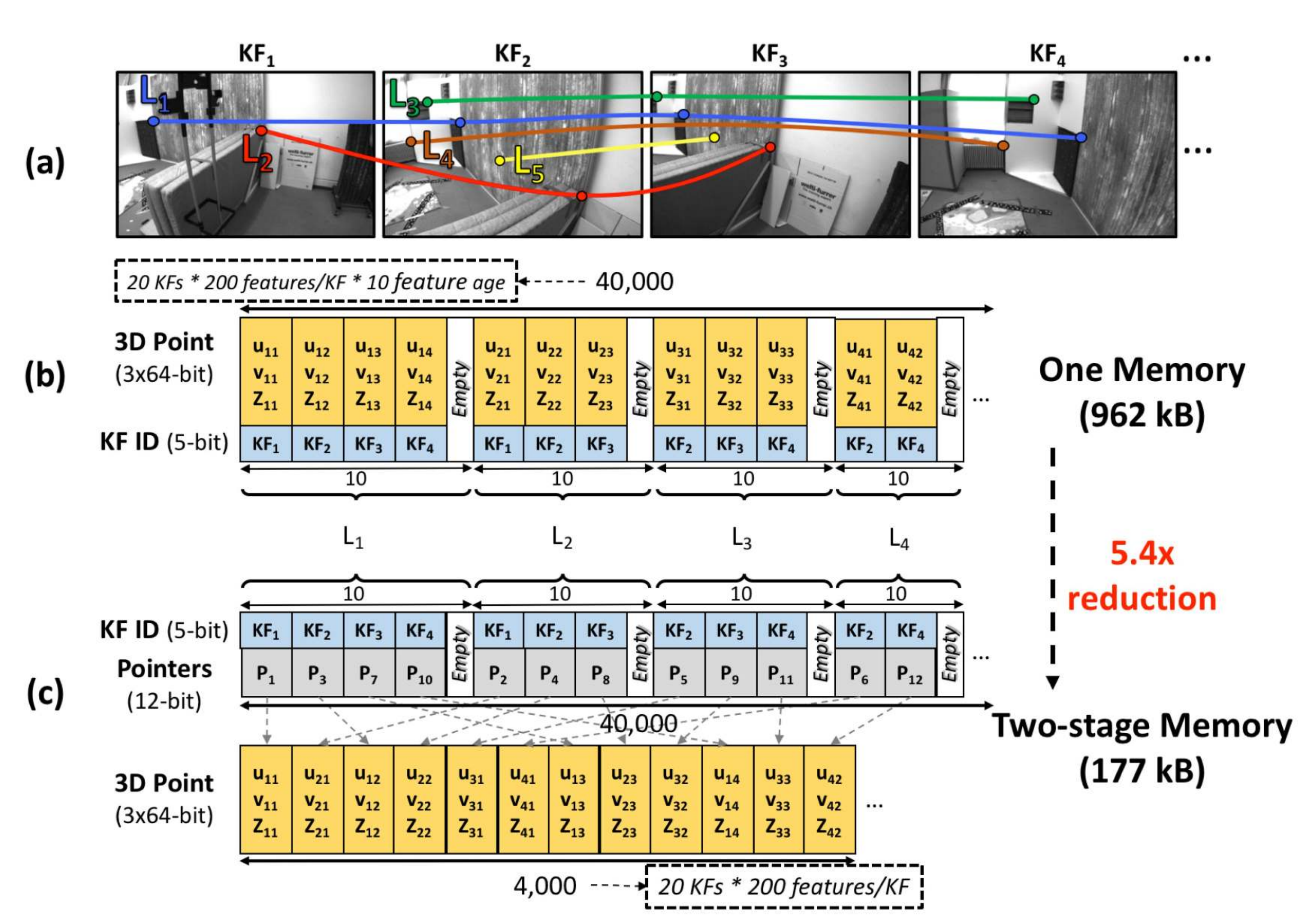}
        \caption{Memory size reduction in the vision factors memory. (a) Examples of feature tracks with variable length. (b) Storing feature tracks in one memory designed for worst case. (c) Dividing the memory into two stages with pointers.}
        \label{fig:vision_factors}
    \end{center}
\end{figure}

Feature tracks account for 88\% of the factor graph memory in $BE$. They contain all observed features in the current horizon, and are used by $BE$ to solve the non--linear optimization problem. Each feature track stores the $KF$ IDs where the landmark is observed in, and the 3D coordinates of the corresponding feature in each $KF$. Fig.~\ref{fig:vision_factors}--a shows an example of some feature tracks. Although a maximum feature age is defined (e.g., 10 in Navion), feature tracks have variable length depending on the image sequence. A feature track can be short because its landmark gets out of the field of view while the camera is moving (e.g., L\textsubscript 2), or because the tracking algorithm fails to track a landmark over time (e.g., L\textsubscript 5).

Fig.~\ref{fig:vision_factors}--b shows the feature tracks stored in one memory. Designing for the worst case, the memory has to store all observations in 20 $KF$s (i.e., maximum number of $KF$s in a horizon) each having 200 features (i.e., maximum number of features tracked per frame) and 10 observations per landmark (i.e., maximum feature age). This results in a big 962kB SRAM, with 40,000 entries, each containing a 5--bit $KF$ ID and 3 64--bit double precision numbers for the 3D coordinates per observation. This memory is populated with measurements from $VFE$, and it is continuously changing with old features being removed and new features being added.

To reduce the size of this large memory, we noticed that it is sparsely populated with a maximum of 4,000 observations from $VFE$ (200 features tracked per frame, 20 $KF$s in a horizon). However, the distribution of these non--zero entries is unknown and depends on the feature track length (Fig.~\ref{fig:vision_factors}--a). As a result, a two--stage memory architecture is used as shown in Fig.~\ref{fig:vision_factors}--c. The first sparse memory still has 40,000 entries, but rather than storing the 3D double precision coordinates, it stores 12--bit pointers to the second dense memory, which stores the 3D values of the 4,000 observations. This reduces the graph memory size by 5.4$\times$, with an overhead of increasing access latency by only one cycle.

\subsection{Linear Solver Structured Sparsity}
\label{sec:structured_sparsity}
\begin{figure}
    \begin{center}
        \includegraphics[width=1.0\linewidth]{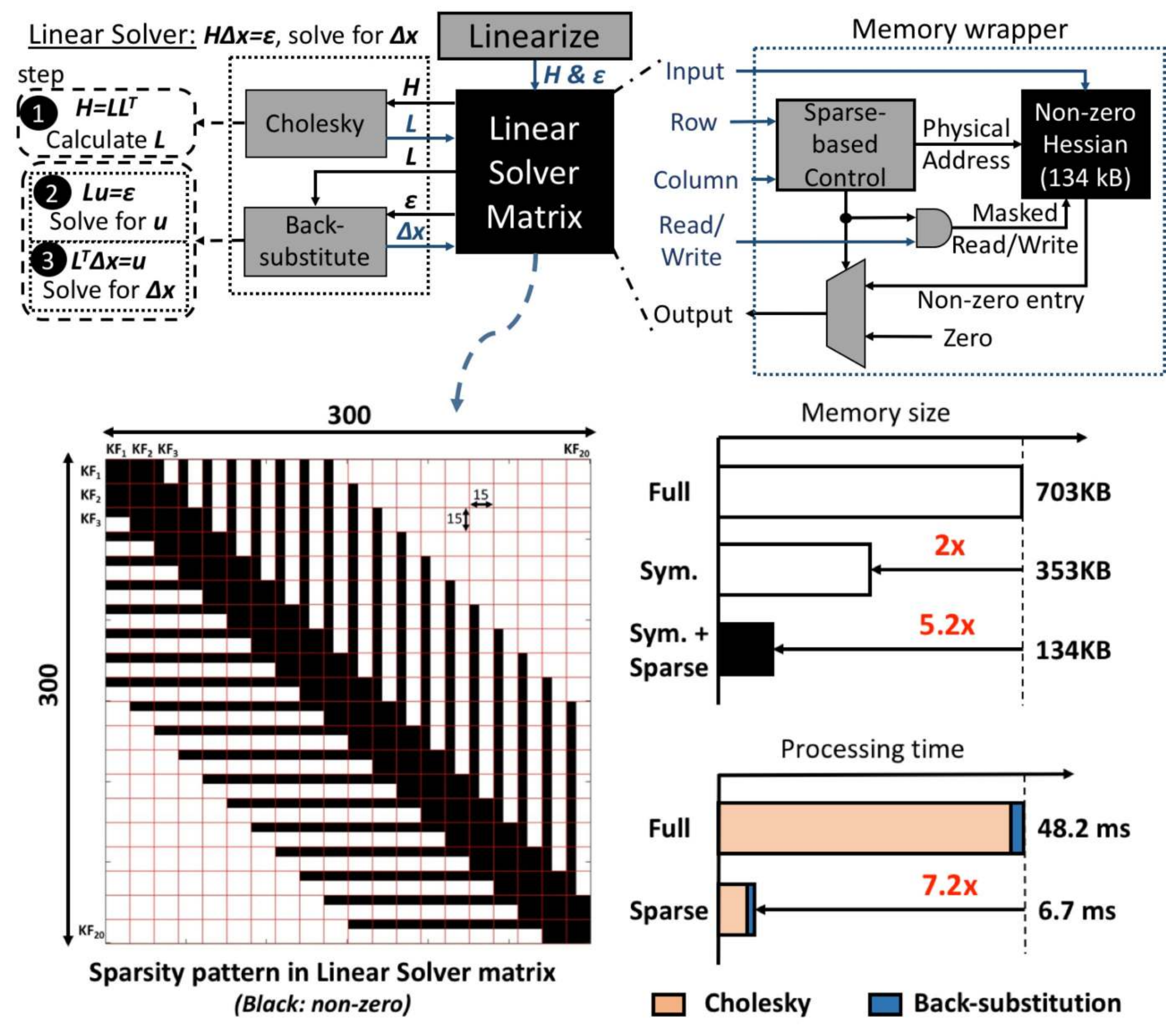}
        \caption{Fixed sparsity visualization in linear solver matrix, with memory size reduction and throughput increase.}
        \label{fig:hessian_sparsity}
    \end{center}
\end{figure}

The first step in $BE$ to solve the non--linear factor graph optimization problem is to linearize all factors in the horizon. The result is a linear system of equations ($H\Delta x = \varepsilon$), where $H$ is the Hessian matrix and $\varepsilon$ is the residual error resulting from the linearization process. With a maximum of 20 $KF$s in the horizon, matrix $H$ size is 300$\times$300 as shown in Fig.~\ref{fig:hessian_sparsity}, where each $KF$'s state contains 15 variables: 3 for position, 3 for orientation, 3 for velocity, and 6 for IMU bias. The $H$ matrix is stored in the linear solver memory in $BE$, where it is updated every $KF$.

The $H$ matrix, shown in Fig.~\ref{fig:hessian_sparsity}, has some characteristics that enable memory size reduction. For instance, $H$ is symmetric, which directly results in 2$\times$ memory size reduction by only storing the upper (or lower) triangle. Additionally, based on the ratio between the feature track age and the horizon size, only 38\% of each of the matrix's triangles have non--zero values, labeled in black in Fig.~\ref{fig:hessian_sparsity}, and their positions are fixed. By storing only the non--zero values, a total 5.2$\times$ memory size reduction is achieved. Fig.~\ref{fig:hessian_sparsity} shows the linear solver memory wrapper. A sparse--based control unit takes the read/write requests with the row and column addresses, and it controls whether to perform the read/write operation on the small 134kB memory or to mask it according to the fixed sparsity pattern.

\begin{figure}
    \begin{center}
        \includegraphics[width=0.9\linewidth]{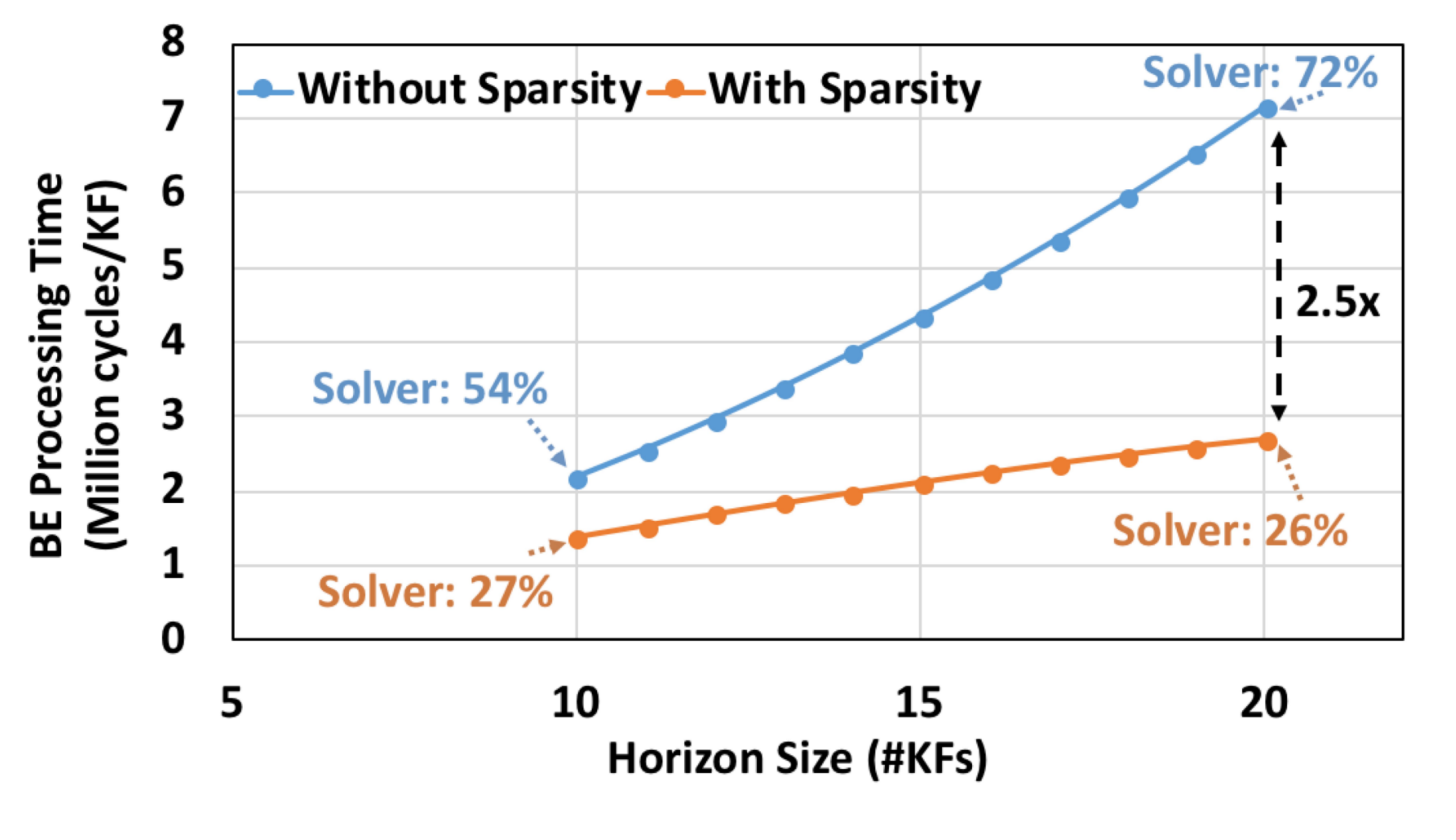}
        \caption{BE processing time savings by exploiting sparsity in the linear solver matrix. The horizon size (in number of KFs) defines the size of the linear system. Percentages show how much time is spent in the linear solver relative to BE processing time.}
        \label{fig:linearSolver_throughput}
    \end{center}
\end{figure}

Fig.~\ref{fig:hessian_sparsity} also shows the linear solver process, happening in--place in the linear solver memory. Cholesky factorization and back--substitution are used to solve $H\Delta x = \varepsilon$. These two operations involve traversing the matrix row by row and column by column. Taking into account the fixed sparsity pattern of $H$, the linear solver processing time can be reduced by skipping processing the zero locations. With the maximum horizon size of 20 $KF$s, a 7.2$\times$ speed--up is achieved by exploiting the $H$ matrix structured sparsity. Fig.~\ref{fig:linearSolver_throughput} shows the $BE$'s overall processing time savings with different horizon sizes. The linear solver processing time increases with $O(N^3)$ without sparsity, compared to an increase much slower than $O(N)$ when exploiting sparsity. This results in more time savings as the number of $KF$s in the horizon increases. At 20 $KF$s in the horizon, a maximum of 2.5$\times$ speed--up of $BE$ processing time is achieved.

\subsection{Rescheduling and Parallelism}
\label{sec:dataflow}
Parallelism is carefully used to increase the throughput with minimal overhead. $VFE$ in particular is suitable for parallelism due its nature of running image processing computation. The main advantage of rescheduling is that it achieves processing time savings without any effect on the overall system accuracy.

\begin{figure*}
    \begin{center}
        \includegraphics[width=0.9\linewidth]{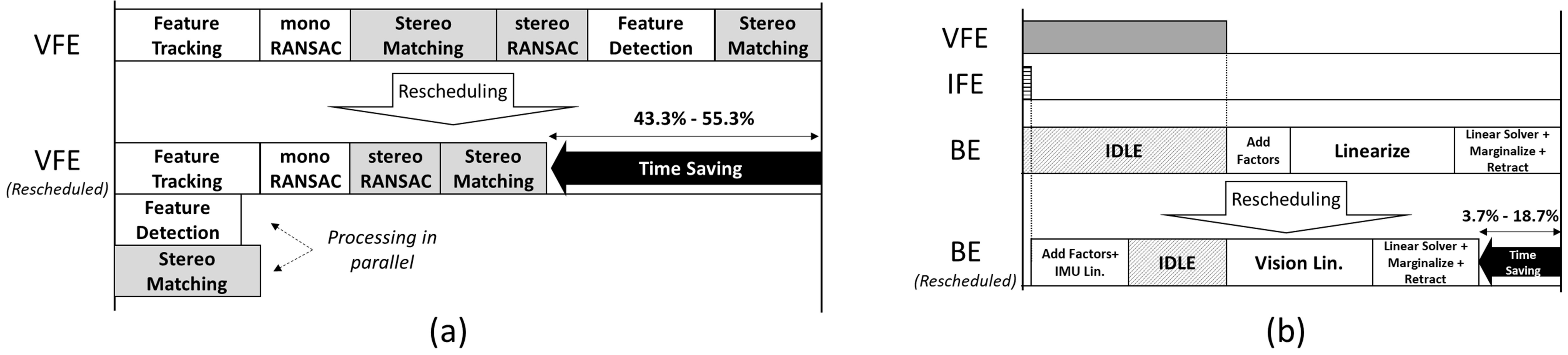}
        \caption{(a) Rescheduling VFE's pipeline processing $KF$s. (b) Rescheduling BE's pipeline.}
        \label{fig:rescedule}
    \end{center}
\end{figure*}

Fig.~\ref{fig:rescedule}--a shows the rescheduling of $VFE$'s pipeline to make use of parallel processing in hardware. Algorithmically, the image processing parts of the $VFE$ can run in parallel. Although the algorithm has some dependency where $FD$ waits for $FT$ and $RANSAC$ to know how many new features are needed to be detected (Fig.~\ref{fig:rescedule}--a top), this dependency can be broken by pre--detecting features and selecting the relevant features later (Fig.~\ref{fig:rescedule}--a bottom). This parallelism is not exploited in software because firing up multiple cores can significantly increase power consumption, which is already high with one core. The rescheduling results in $VFE$'s processing time saving between 43\% to 55\% depending on the environment. This comes with a 77kB (10\%) memory overhead with line buffers to support the required bandwidth of the parallel components.

Another rescheduling is carried out in $BE$'s pipeline to enable parallelism between $BE$ and $VFE$. At $KF$s, $BE$ waits for $VFE$ and $IFE$ to process their outputs. However, $IFE$ is much faster and its output is ready earlier than $VFE$ as shown in Fig.~\ref{fig:rescedule}--b. $BE$ processing can be rescheduled such that the initialization and all IMU factors linearization start immediately after $IFE$'s output is available. This results in $BE$'s processing time saving between 4\% to 19\% depending on the scene, with no overhead.

\section{Implementation and Results}
\label{sec:results}

\subsection{Chip Implementation Results}
\begin{figure}
    \begin{center}
        \includegraphics[width=0.95\linewidth]{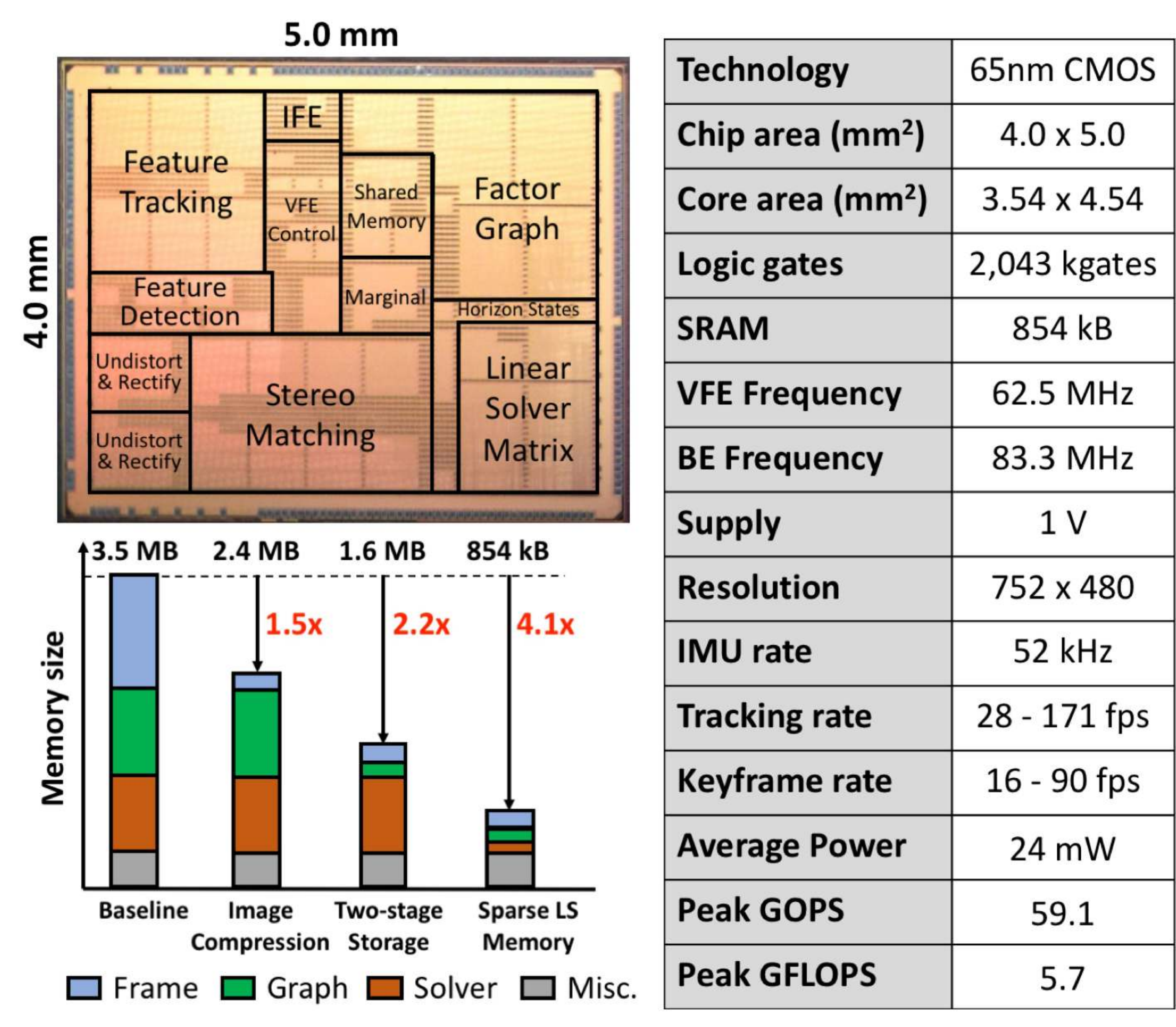}
        \caption{Die photo and summary of the chip specifications.}
        %\vspace{-10pt}                
        \label{fig:die_specs}
    \end{center}
\end{figure}

Fig.~\ref{fig:die_specs} shows the die photo of Navion chip with a summary of the chip specifications and the memory optimization results. Navion is implemented in a 65nm CMOS technology with 2 million NAND2 equivalent logic gate count and 854kB on--chip SRAM. The chip contains two clock domains, one for $VFE$ and the other for both $IFE$ and $BE$. By using image compression and exploiting structured and unstructured sparsity, an overall 4.1$\times$ memory saving is achieved, which enables full $VIO$ pipeline integration on--chip. Table~\ref{tab:memories} shows different memory sizes before and after optimizations.

\begin{table}
    \centering
    \begin{tabular}{|l|l|c|c|c|}
         \hline 
         & \multirow{2}{4em}{\textbf{Memory Blocks}} & \multicolumn{2}{|c|}{\textbf{Size (kB)}} & \multirow{2}{4em}{\textbf{Savings}} \\ \cline{3-4}
          & & Before & After & \\ \hline
          \multirow{5}{3em}{\textbf{VFE}} & Frame Buffers & 1410 & 317.6 & 4.4$\times$ \\ \cline{2-5}
          & Line Buffers & \multicolumn{2}{|c|}{78.2} & -- \\ \cline{2-5}
          & Tracking Data & \multicolumn{2}{|c|}{18.2} & -- \\ \cline{2-5}
          & Features Candidates & \multicolumn{2}{|c|}{16} & -- \\ \cline{2-5}
          & Keyframe Data & \multicolumn{2}{|c|}{11.7} & -- \\ \hline
          \textbf{IFE} & Integrated Data & \multicolumn{2}{|c|}{2} & -- \\ \hline
          \multirow{4}{3em}{\textbf{BE}} & Factor Graph & 1012 & 205.6 & 4.9$\times$ \\ \cline{2-5}
          & Linear Solver Data & 882.4 & 163.5 & 5.4$\times$ \\ \cline{2-5}
          & Horizon States & \multicolumn{2}{|c|}{5.2} & -- \\ \cline{2-5}
          & Shared Memory & \multicolumn{2}{|c|}{36} & -- \\ \hline
         \multicolumn{2}{|c|}{\textbf{Total}} & \textbf{3471.7} & \textbf{854} & \textbf{4.1$\times$} \\ 
         \hline
    \end{tabular}
    \caption{The size of different memory blocks in Navion, showing numbers before and after optimizations in Section~\ref{sec:optimizations}.}
    \label{tab:memories}
\end{table}

Navion can process stereo images with a maximum resolution of 752$\times$480 at a rate of 28--171 fps in real--time across the different sequences in EuRoC dataset~\cite{roboticss2016Burri} used for evaluation; this is referred to as the tracking rate. The chip can also process inertial measurements at up to 52kHz. Navion updates the states and the sparse 3D map at $KF$ rate of 16--90 fps, which is the $BE$ rate, also depending on the sequence. It consumes an average power consumption of 24mW when operating at 1V. These numbers are measured when Navion's programmable parameters are set to their maximum values; the main parameters are shown in Table~\ref{tab:max_params}. Changing these parameters values can trade--off accuracy, throughput and power consumption across different environments, as shown in Section~\ref{sec:adapt}.

\begin{table}
\centering
    \begin{tabular}{|l|c|c|}
        \hline
         \multicolumn{2}{|l|}{\textbf{Parameter}} & \textbf{Maximum Value} \\ \hline
          \multicolumn{2}{|l|}{Frame resolution} & 752$\times$480 \\ \hline
         \multicolumn{2}{|l|}{Number of features per frame} & 200 \\ \hline
         \multirow{3}{4em}{Feature Tracking} & Pyramid (P\textsubscript{FT}) & 3 levels \\ \cline{2-3}
                                             & Cell size (C\textsubscript{FT}) & 15$\times$ 15 pixels \\ \cline{2-3}
                                             & Iterations (I\textsubscript{FT}) & 30 per level \\ \hline
         \multirow{2}{4em}{Stereo} & Template size (T\textsubscript{S}) & 51$\times$ 5 pixels \\ \cline{2-3}
                                   & Search region (R\textsubscript{S}) & 421$\times$ 5 pixels \\ \hline
         \multirow{3}{4em}{Backend} & Horizon size & 20 Keyframes\\ \cline{2-3}
                                   & Feature age & 10 Keyframes\\ \cline{2-3}
                                   & Feature tracks & 4000 tracks \\
         \hline
    \end{tabular}
    \caption{Main programmable parameters with their maximum values. }
    \label{tab:max_params}
\end{table}

\begin{figure}
    \begin{center}
        \includegraphics[width=0.90\linewidth]{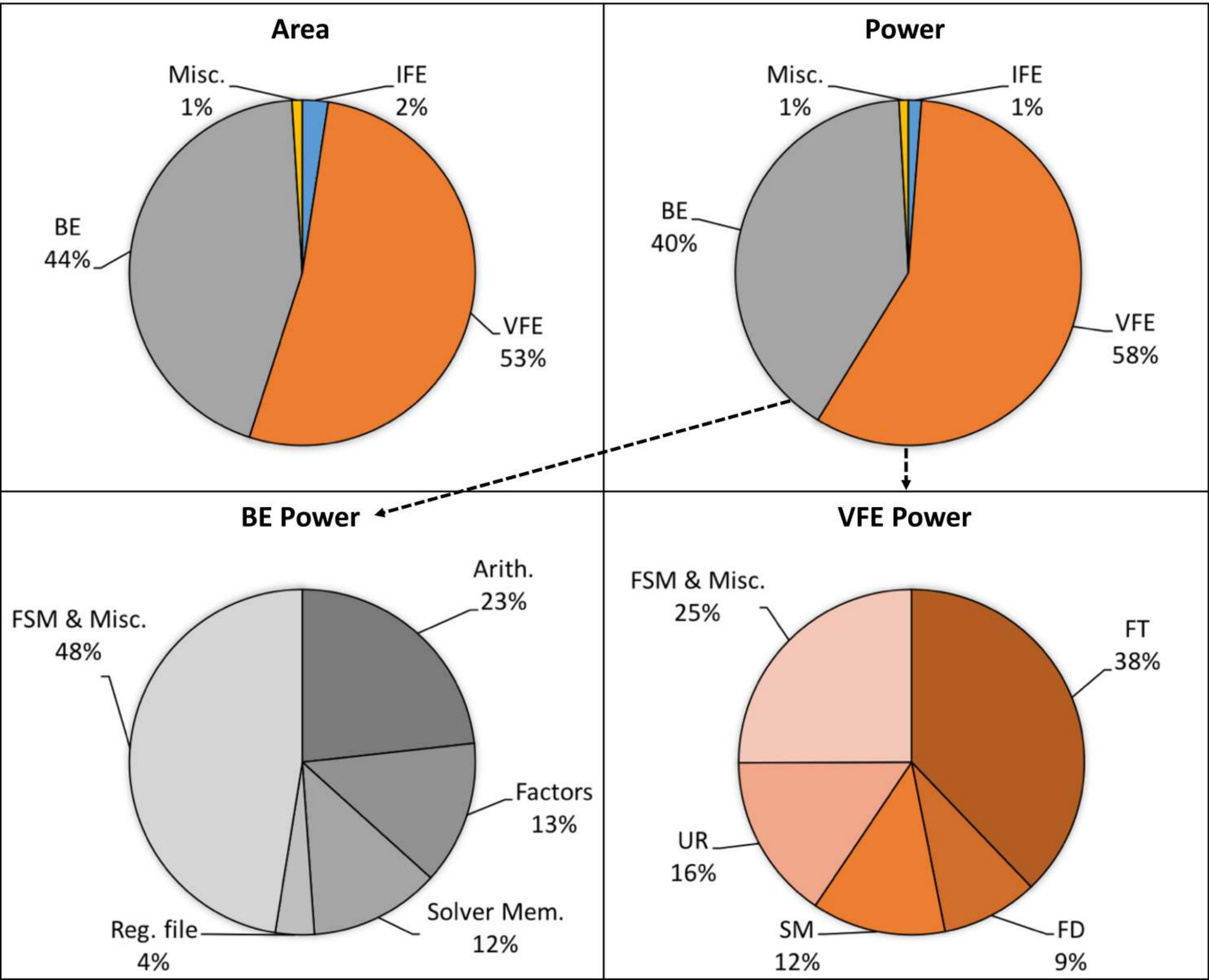}
        \caption{Area and power breakdown.}
        %\vspace{-10pt}                
        \label{fig:navion_breakdown}
    \end{center}
\end{figure}

Fig.~\ref{fig:navion_breakdown} shows Navion's area and power breakdowns. $IFE$ has a very low cost from both area and power points of view, while $VFE$ and $BE$ have relatively similar area. The average power consumption of $BE$ is 40\% of the total power since it is active only at $KF$s. It's worth--mentioning that the large on--chip SRAM power is only 21\% of the total power consumption, although it accounts for 77\% of the chip area. This is a result of dividing the large memories (i.e., frame buffers, vision factors memory, etc.) into small banks, and clock--gating the ones that are not used. This results in 5--9$\times$ power savings, depending on the memory size.

Fig.~\ref{fig:navion_breakdown} also shows a detailed power breakdown of $VFE$ and $BE$, where different dataflows result in different power distribution. In $VFE$, parallel image processing modules (i.e., $FT$, $FD$, $UR$, and $SM$) consume the majority of the power compared to control and $FSM$. $FT$ consumes the largest amount of power in $VF$E because it is always active regardless of the frame type (i.e., $KF$ or not). However, in $BE$, half of the power is consumed by the control and $FSM$ because of the serial nature of the $BE$'s architecture. Additionally, almost one quarter of $BE$'s power is consumed by the shared arithmetic unit.

\begin{figure}
    \begin{center}
        \includegraphics[width=1.0\linewidth]{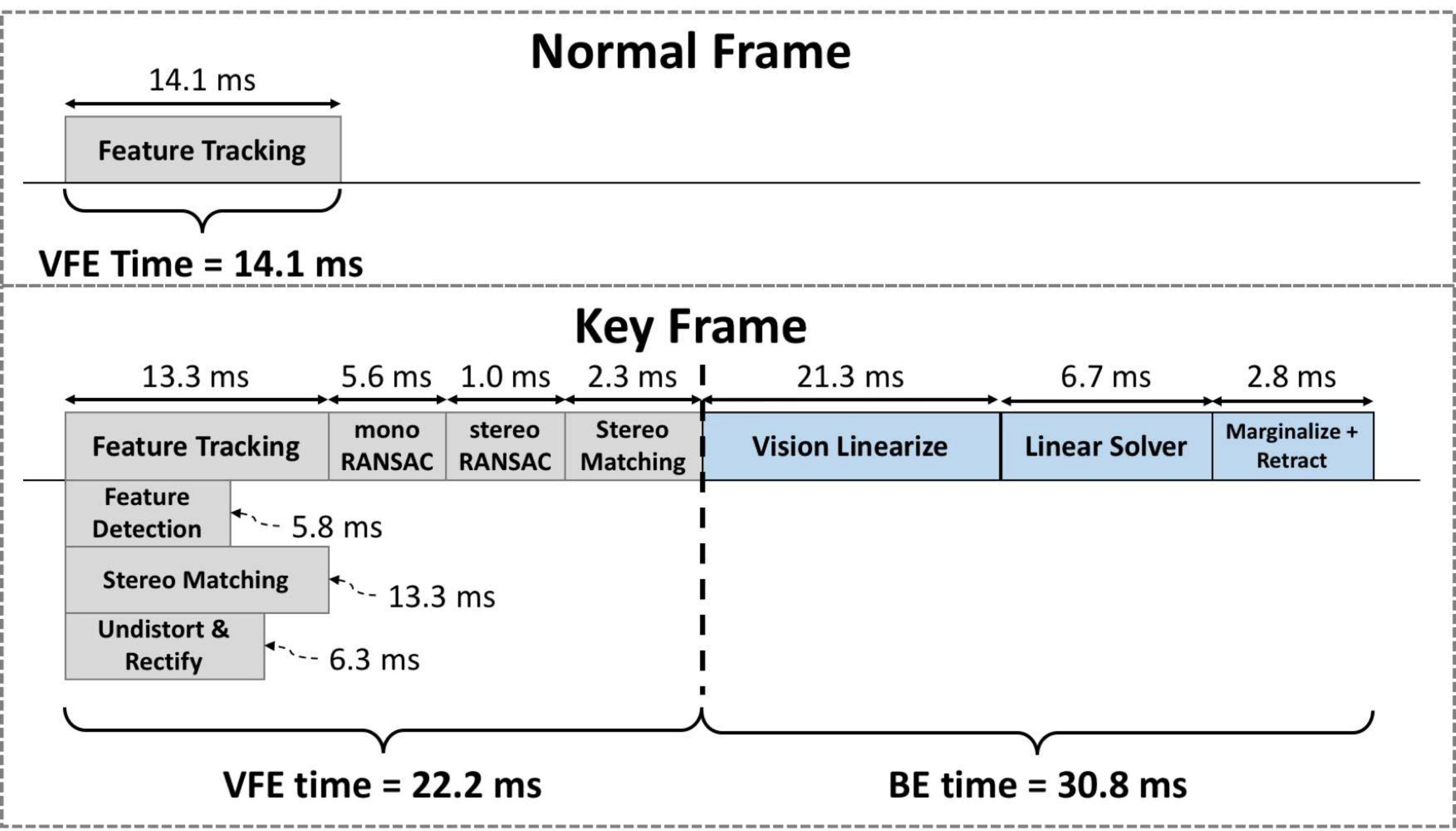}
        \caption{Navion's timing breakdown at maximum configuration (62.5 MHz VFE clock and 83.3 MHz BE clock). Numbers are averages over EuRoC dataset.}
        %\vspace{-10pt}                
        \label{fig:timing_breakdown}
    \end{center}
\end{figure}

Fig.~\ref{fig:timing_breakdown} shows a detailed timing breakdown for the different VIO modules, in both non--$KF$ and $KF$, measured on average over different sequences in EuRoC dataset (see Section~\ref{sec:evaluation}). Based on the rescheduling optimization discussed in Section~\ref{sec:dataflow}, $FD$, $UR$, and $SM$ modules all run in parallel with $FT$ at $KF$s, where their processing time is hidden. For $BE$, the majority of the processing time (65\%) is consumed in linearizing the vision factors. Linearizing IMU and the other factors (not shown in Fig.~\ref{fig:timing_breakdown}) is carried out in parallel with $VFE$ processing, and it takes less than 2 ms to finish. Additionally, the linear solver is relatively fast (i.e., only 25\% of $BE$'s processing time) as it exploits sparsity by skipping processing of zeros.

\subsection{Demonstration System}
\label{sec:demo}
\begin{figure*}
    \begin{center}
        \includegraphics[width=0.90\linewidth]{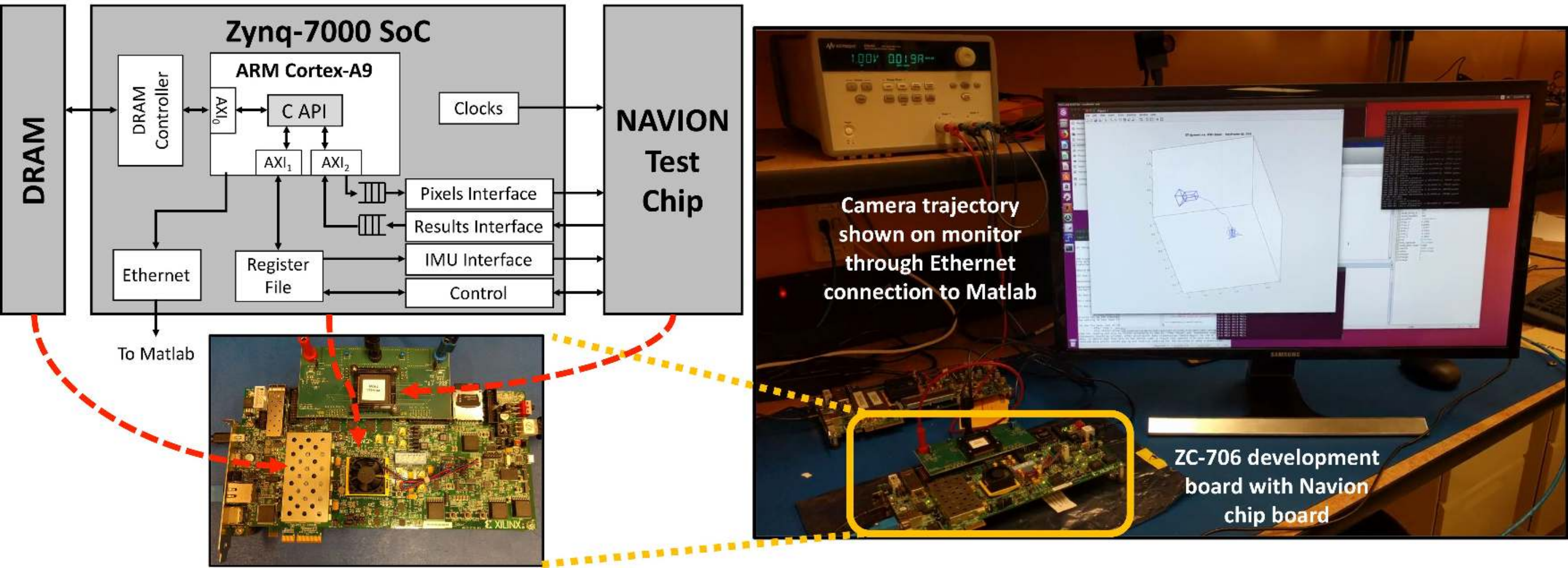}
        \caption{Demonstration system. ARM core in Zynq--7000 SoC is connected to the on--board DRAM thought AXI0 bus. AXI1 and AXI2 buses are used to connect the ARM core to the control and the interface logic on the FPGA fabric. The two clock signals needed to run the chip are generated on the FPGA. }
        %\vspace{-10pt}                
        \label{fig:demo}
    \end{center}
\end{figure*}

To validate the fabricated chip, a demonstration system is developed for real--time localization and mapping as shown in Fig.~\ref{fig:demo}. It is composed of the custom test chip board and a Xilinx ZC-706 evaluation board. The FPGA board is used to stream images and IMU measurements to the chip, and read the output results for verification and visualization. The Xilinx Zynq--7000 FPGA has 2 ARM Cortex--A9 embedded cores. A C API is developed on an Ubuntu operating system in the ARM core to control the dataflow through several buses using AXI protocol. Fig.~\ref{fig:demo} shows Navion's output trajectory drawn on the monitor in real--time. A video of this demo system can be found on the Navion project website~\cite{navion_website}. 

%AXI0 bus is connected to the on-board DRAM through a controller IP to fetch the images and IMU measurements. AXI1 and AXI2 buses are used to connect the ARM core to the control and the interface logic on the FPGA fabric. AXI1 bus is connected to a register file which is mainly used for control signals, in addition to streaming IMU measurements because of their low bandwidth. AXI2 bus is connected to two FIFOs; one to send the stereo images to Navion, and the other to read the results from Navion. Finally, the two clock signals needed to run the chip are generated on the FPGA. 
%The data is fed to the system through DRAM under the Ubuntu operating system environment, but practically this interface can be directly connected to camera and IMU sensors

\subsection{Evaluation Results}
\label{sec:evaluation}
To evaluate Navion's accuracy, we use the EuRoC dataset~\cite{roboticss2016Burri} which is one of the most challenging and widely used datasets for UAVs flying indoors. EuRoC dataset contains 11 different sequences, each representing the UAV flying in one of three different rooms: \textit{Machine Hall} ($MH$), \textit{Vicon room 1} ($V1$), and \textit{Vicon room 2} ($V2$). Some of the sequences are quite challenging as they have relatively fast and unstable motion and strong brightness change, which lead to dark and/or blurred images. The sequences are divided into easy, medium and difficult sequences accordingly.

%Some of the sequences are quite challenging as they have relatively fast and unstable motion and strong brightness change as shown in Fig.~\ref{fig:euroc_examples}, which lead to dark (Fig.~\ref{fig:euroc_examples}-b) or blurred (Fig.~\ref{fig:euroc_examples}-d) images. The sequences are divided into easy, medium and difficult sequences accordingly.

%\begin{figure}
%    \begin{center}
%        \includegraphics[width=1.0\linewidth]{figures/Fig_euroc_dataset.pdf}
%        \caption{Examples from EuRoC dataset.}
%        %\vspace{-10pt}                
%        \label{fig:euroc_examples}
%    \end{center}
%\end{figure}

\begin{figure}
    \begin{center}
        \includegraphics[width=0.75\linewidth]{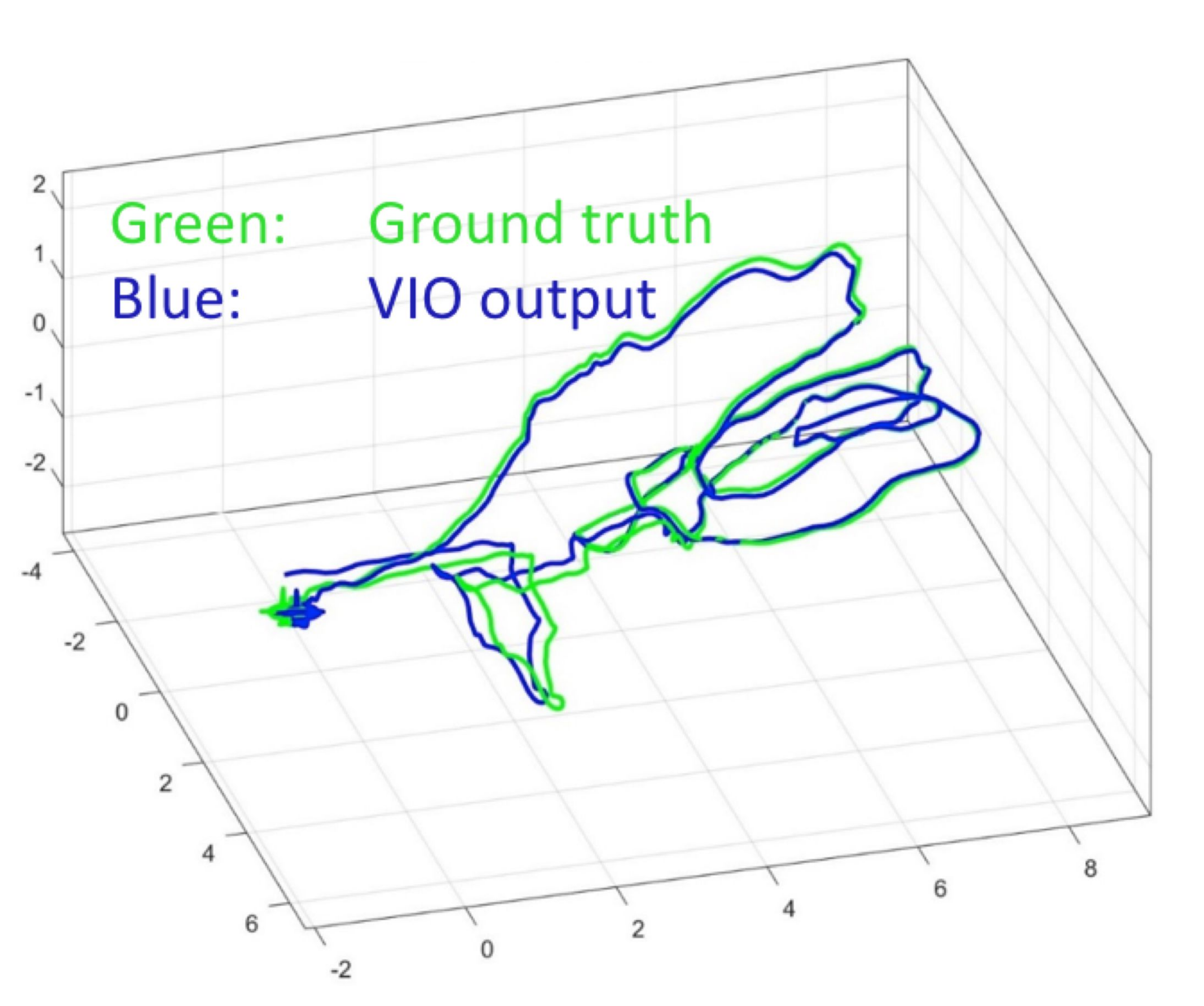}
        \caption{VIO output example with ground truth, processing a sequence from EuRoC dataset.}
        %\vspace{-10pt}                
        \label{fig:trajectory_output}
    \end{center}
\end{figure}

Table~\ref{tab:basline_performance_summary} shows the average error, throughput and power numbers of the VIO comparing Navion to software implementations on a desktop Intel Xeon E5-2667 CPU and an embedded ARM Cortex-A15 CPU. The trajectory error is defined as the difference between the VIO output and the ground truth along the whole flight path (Fig.~\ref{fig:trajectory_output}). To account for different sequence lengths, the trajectory error is normalized to the flight length. In this experiment, all sequences are processed with the same configuration parameters in both the software implementation and Navion. Due to randomness in $RANSAC$, all accuracy numbers (i.e., trajectory error) are the average of 5 runs per sequence.

Navion's average trajectory error increases by 6.27 cm, over an average flight length of 83 m in EuRoC dataset. Hence, the normalized error increases from 0.22\% to 0.28\%. This 0.06 percentage point error increase is mainly due to lossy image compression and fixed point arithmetic in $VFE$. Based on the odometry survey in~\cite{robotmag2012Fraundorfer}, different algorithms have relative trajectory errors ranging from 0.1 to 2\%. Thus, Navion has a relatively low average trajectory error compared to these works. Additionally, Navion achieves three orders of magnitude less energy consumption compared to both Xeon and ARM cores.

\begin{table}
\centering
    \begin{tabular}{|l|l|c|c|c|}
        \hline
        \multicolumn{2}{|l|}{\textbf{Platform}}& \textbf{Xeon} & \textbf{ARM} & \textbf{Navion} \\ \hline
        \multicolumn{2}{|l|}{\textbf{Trajectory error (cm)}} & \multicolumn{2}{|c|}{16.98} & 23.25 \\ \hline 
        \multicolumn{2}{|l|}{\textbf{Trajectory error (\%)}} & \multicolumn{2}{|c|}{0.22\%} & 0.28\% \\ \hline 
        \multicolumn{2}{|l|}{\textbf{Average tracking throughput (fps)}} & 63 & 19 & 71 \\ \hline
        
        \multirow{3}{8em}{\textbf{Average $KF$ throughput (fps)}} & \textbf{Frontend} & 19 & 3 & 45 \\ \cline{2-5}
        & \textbf{Backend} & 35 & 6 & 32 \\ \cline{2-5}
        & \textbf{Total} & 12 & 2 & 19 \\ \hline 
        \multicolumn{2}{|l|}{\textbf{Average power (W)}} & 27.9 & 2.4 & 0.024 \\ \hline
        \multicolumn{2}{|l|}{\textbf{Energy (mJ/KF)}} & 3638.6 & 1573.2 & 2.3 \\
        \hline
    \end{tabular}
    \caption{VIO average error, throughput and power processing EuRoC dataset.}
    \label{tab:basline_performance_summary}
\end{table}

\begin{figure}
    \begin{center}
        \includegraphics[width=0.95\linewidth]{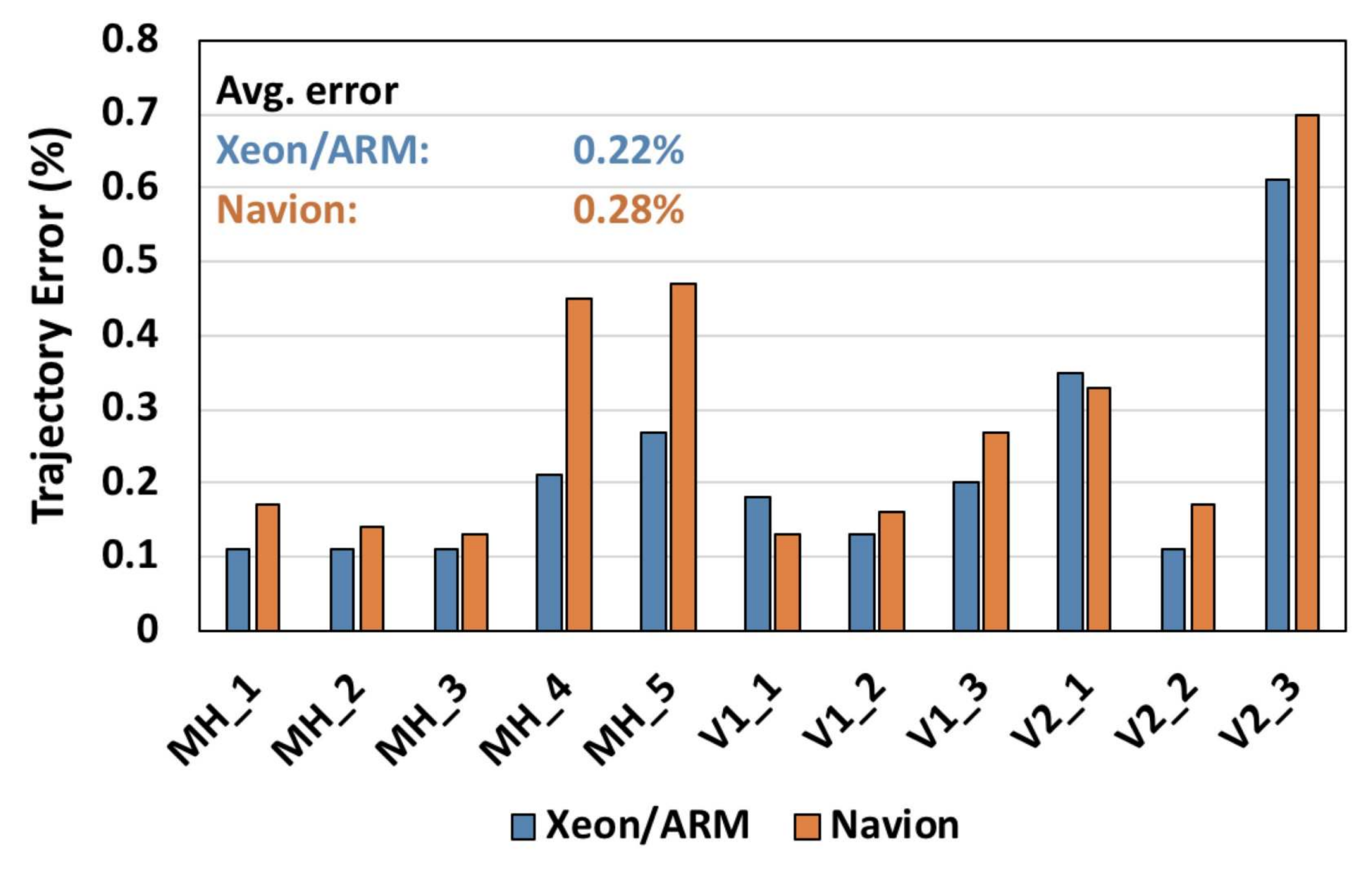}
        \caption{VIO average error for the 11 sequences in EuRoC dataset. VIO parameters are set to their maximum values for all sequences.}           
        \label{fig:baseline_error}
    \end{center}
\end{figure}

Fig.~\ref{fig:baseline_error} shows the VIO trajectory error detailed for the 11 sequences in EuRoC dataset, in both Xeon/ARM CPU and in Navion. As expected, with slow motion, bright scene, and highly textured regions in the easy sequences (i.e., MH\_1, MH\_2, V1\_1 and V2\_1), low trajectory error are achieved in both CPU and Navion. However, the error increases for the blurry, dark, and fast moving medium and difficult sequences. Note that for this experiment, the VIO parameters are set to their maximum settings, in both Navion and the software implementations, for all 11 sequences.

%Comparing Navion to CPU numbers, the fixed point arithmetic and image compression result in a relatively higher trajectory error in Navion for some sequences; like MH\_{4} and MH\_{5}. For the remaining sequences, Navion's error is comparable and sometimes even better than CPU; like V1\_{1} and V2\_{1}. Note that all sequences satisfies the target trajectory error (dashed red line) except V2\_{3}, in both CPU and Navion. Section~\ref{sec_6:adapt} discusses how to reduce the error by adapting to different environments.

\subsection{Adapting to the environment}
\label{sec:adapt}
The flexibility that Navion presents with its programmable parameters gives an opportunity to trade--off throughput, accuracy and power consumption. This is done by configuring the chip differently based on the environment and/or the UAV movement. Accuracy numbers for each sequence, previously shown in Fig.~\ref{fig:baseline_error}, show a large gap between the trajectory error of easy and difficult sequences when all have the same configuration. Particularly, difficult sequences MH\_4, MH\_5 and V2\_3 have relatively large trajectory error when using the same configuration as other sequences. Additionally, adapting to the environment also means that Navion's workload can change based on the environment, resulting in energy savings.

To show the potential savings that can be achieved with adaptation, we set a target normalized trajectory error of 0.35\% for all sequences, and find the optimum configuration for each one independently to meet this target. Table~\ref{tab:adapt_config_error} shows the main adapted parameters for each sequence and the resulting normalized trajectory error. Small number of features and small horizon size are sufficient in easy sequences because it is easy to track them. This reduces the workload of easy sequences, which reduces the energy consumption. More features and longer horizons are used in difficult sequences to account for the short feature tracks.

\begin{table}
\centering
    \begin{tabular}{|l|c|c|c|}
        \hline
         \multirow{2}{3em}{\textbf{Name}} & \multicolumn{2}{|c|}{\textbf{Parameters}} & \multirow{2}{3em}{\textbf{Error}} \\ \cline{2-3}
         & Features/frame & Horizon size & \\ \hline
         MH\_1 & 35 & 10 & 0.33\% \\ \hline
         MH\_2 & 35 & 10 & 0.22\% \\ \hline
         MH\_3 & 35 & 10 & 0.22\% \\ \hline
         MH\_4 & 150 & 10 & 0.29\% \\ \hline
         MH\_5 & 100 & 15 & 0.30\% \\ \hline
         V1\_1 & 35 & 10 & 0.21\% \\ \hline
         V1\_2 & 35 & 10 & 0.19\% \\ \hline
         V1\_3 & 35 & 10 & 0.33\% \\ \hline
         V2\_1 & 50 & 10 & 0.33\% \\ \hline
         V2\_2 & 35 & 10 & 0.16\% \\ \hline
         V2\_3 & 50 & 15 & 0.34\% \\ 
         \hline
    \end{tabular}
    \caption{VIO independent parameters per sequence for 0.35\% target trajectory error.}
    \label{tab:adapt_config_error}
\end{table}

\begin{figure}
    \begin{center}
        \includegraphics[width=1.0\linewidth]{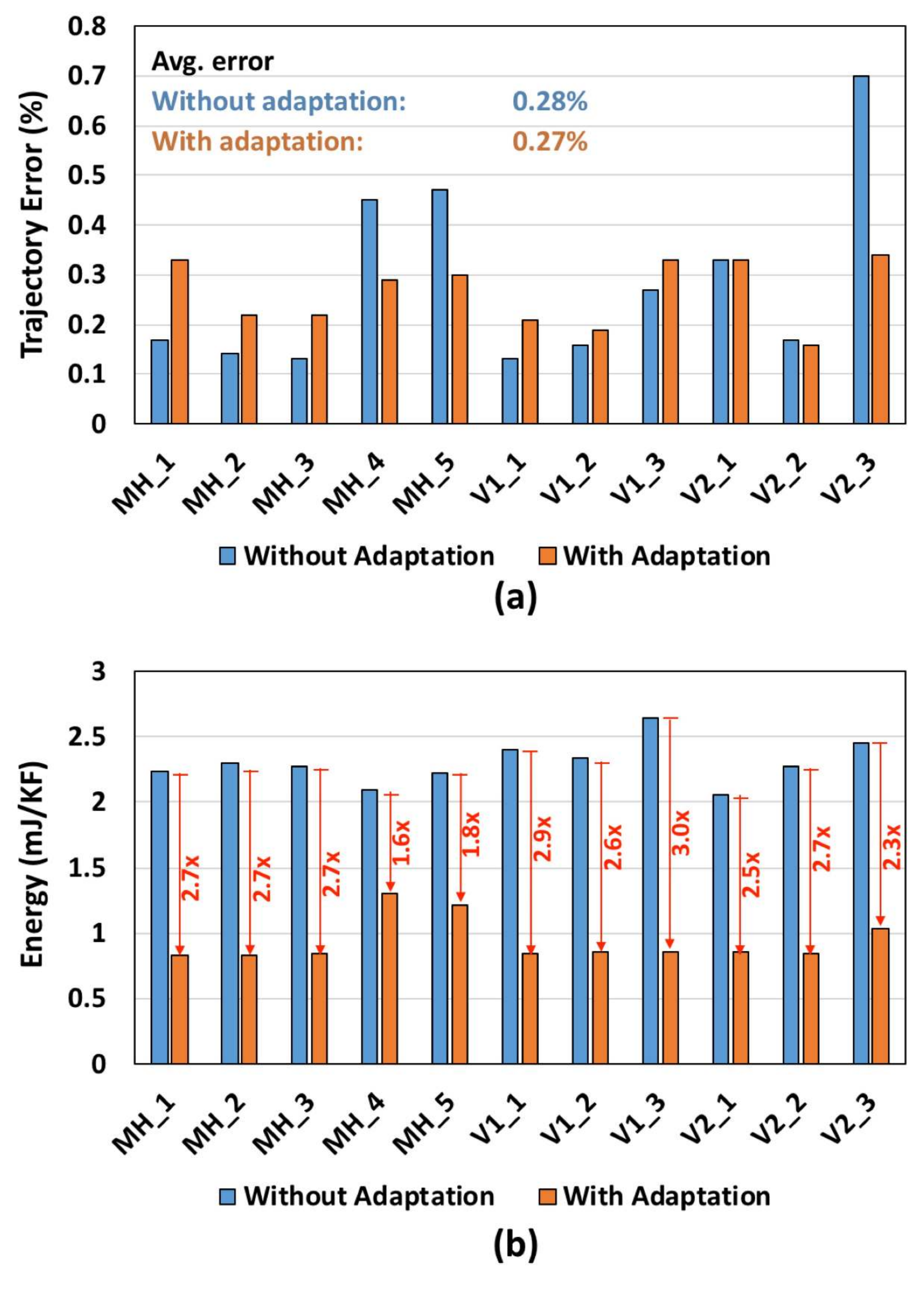}
        \vspace{-10pt}
        \caption{With and without adaptation comparisons for the 11 sequences in EuRoC dataset. (a) Trajectory error. (b) Energy consumption.}
        \label{fig:adapt_energy}
    \end{center}
\end{figure}

Fig.~\ref{fig:adapt_energy} shows a comparison between Navion's measurements with and without adaptation. Fig.~\ref{fig:adapt_energy}--a shows the trajectory error for each sequence. Although the average trajectory error remains the same for both cases, the individual error numbers for the easy sequences (i.e., MH\_1, MH\_2, V1\_1 and V2\_1) increased slightly but are still within the 0.35\% target, and the error for difficult sequences (i.e., MH\_4, MH\_5 and V2\_3) decreased significantly. Fig.~\ref{fig:adapt_energy}--b shows the energy consumption for each sequence. The energy numbers for all sequences decreased significantly by an average of 2.5$\times$.

Table~\ref{tab:different_config} shows the average throughput and energy numbers when adapting Navion for each EuRoC dataset sequence. Different throughput modes are shown such that: 1) Maximum rate means that Navion is running at its maximum frequency for all sequences. 2) Fixed distance means that the throughput of each sequence is set for a constant distance between $KF$s (5 cm/$KF$). 3) Fixed rate means that Navion is running at the rate of which EuRoC dataset is captured (i.e, 20 fps). In this third configuration, and with adaptation, Navion can process the EuRoC dataset while consuming only an average power of 2mW. This experiment is clearly showing the importance of having a configurable VIO accelerator that can adapt to the environment, not only for an improved accuracy, but also for large energy and power savings. In all these experiments, the power supply was set to 1V. Additional power and energy savings can be achieved by lowering down the supply voltage.

% \begin{table*}
% \centering
%     \begin{tabular}{|l|c|c|c|c|c|c|}
%         \hline
%         & \multicolumn{3}{|c|}{\textbf{Without Adaptation}} & \multicolumn{3}{|c|}{\textbf{With Adaptation}} \\ \hline
        
%         \multirow{2}{10em}{\textbf{Throughput Mode}} & Maximum & Constant & Constant & Maximum & Constant & Constant \\
        
%         & Throughput & Distance & Throughput & Throughput & Distance & Throughput \\ \hline
        
%         \textbf{Trajectory Error} & \multicolumn{3}{|c|}{0.28\%} & \multicolumn{3}{|c|}{0.27\%} \\ \hline
        
%         \textbf{Average Tracking (fps)} & 71 & 41 & 20 & 142 & 41 & 20 \\ \hline
        
%         \textbf{Average KF (fps)} & 31 & 14 & 5 & 100 & 14 & 5 \\ \hline
        
%         \textbf{Average Power (mW)} & \textbf{24} & \textbf{18} & \textbf{6} & \textbf{22} & \textbf{7} & \textbf{2} \\ 
%         \hline
%     \end{tabular}
%     \caption{Navion  average error, thought and power numbers for different configurations, with and without adaptation.}
%     \label{tab:different_config}
% \end{table*}

\begin{table}
\centering
    \begin{tabular}{|l|c|c|c|c|c|c|}
        \hline
        & \multicolumn{3}{|c|}{\textbf{Without Adaptation}} & \multicolumn{3}{|c|}{\textbf{With Adaptation}} \\ \hline
        
        \multirow{2}{3em}{\textbf{Configuration}} & Max & Fixed  & Fixed & Max & Fixed & Fixed \\
        
        & Rate & Dist & Rate & Rate & Dist. & Rate \\ \hline
        
        \textbf{Traj. Error} & \multicolumn{3}{|c|}{0.28\%} & \multicolumn{3}{|c|}{0.27\%} \\ \hline
        
        \textbf{VFE Clock (MHz)} & 62.5 & 45.7 & 16.8 & 62.5 & 18.6 & 6.6 \\ \hline
        
        \textbf{BE Clock (MHz)} & 83.3 & 60.9 & 22.5 & 83.3 & 24.8 & 8.8 \\ \hline   
        
        \textbf{Tracking (fps)} & 71 & 41 & 20 & 142 & 41 & 20 \\ \hline
        
        \textbf{KF (fps)} & 19 & 14 & 5 & 49 & 14 & 5 \\ \hline
        
        \textbf{Power (mW)} & \textbf{24} & \textbf{18} & \textbf{6} & \textbf{22} & \textbf{7} & \textbf{2} \\ 
        \hline
    \end{tabular}
    \caption{Navion's error, throughput, clock frequencies and power numbers for different configurations, with and without adaptation. Numbers are averages over EuRoC dataset.}
    \label{tab:different_config}
\end{table}
%\input{06_adapt}
%!TEX root=main.tex
\vspace{-5pt}
\section{Conclusions}
\label{conclusion}

Scaling down localization and mapping to nano and pico drones/robots requires hardware and algorithm co--design. In this paper, we improve our initial co--designed visual--inertial odometry on FPGA and propose the first fully integrated ASIC solution, Navion. A simple yet effective on--chip image compression technique is developed to reduce the size of the memory. A specialized memory architecture is proposed to efficiently store the mono/stereo feature tracks within the whole horizon. Both structured and unstructured sparsity patterns of the memory for the $BE$ are exploited to further reduce the memory and increase the throughput.

Navion is fabricated in a 65nm CMOS technology. Several important parameters can have significant impact on the trade--off in throughput, accuracy and energy efficiency of Navion. These parameters include, for example, keyframe rate, horizon size, and number of feature tracks. They are designed to be programmable, allowing the chip to prioritize different goals under different scenarios. At its peak performance, Navion can process 752$\times$480 stereo image at up to 171 fps and inertial measurements at up to 52 kHz while consuming an average of 24mW at 1V. When configured to process EuRoC dataset at the sensor rate of 20 fps, the chip consumes only an average of 2mW at 1V. Thus, Navion can adapt to handle different environments while being energy efficient and processing in real--time, which makes it suitable for different applications such as autonomous navigation, mapping, and portable AR/VR.
%\section*{Acknowledgments} The authors would like to thank the TSMC University Shuttle Program for chip fabrication and NSF for providing funding.

%\printbibliography[title=Bibliography]

\bibliographystyle{unsrt}
\small
%\bibliography{refs}

\begin{thebibliography}{10}

\bibitem{geomatics2014}
Francesco Nex and Fabio Remondino.
\newblock {UAV for 3D mapping applications: a review}.
\newblock {\em Applied Geomatics}, 6(1):1--15, 2014.

\bibitem{icuas2014Mohammed}
F~Mohammed, A~Idries, N~Mohamed, K~Al-Jaroodi, and I~Jawhar.
\newblock {UAVs for smart cities: Opportunities and challenges}.
\newblock In {\em International Conference on Unmanned Aircraft Systems (ICUAS)}, 2014.

\bibitem{uav2015Valavanis}
Kimon~P Valavanis and George~J Vachtsevanos.
\newblock {UAV Applications: Introduction}.
\newblock In {\em Handbook of Unmanned Aerial Vehicles}, pages 2639--2641. Springer Netherlands, Dordrecht, 2015.

\bibitem{access2017Chatzopoulos}
D.~Chatzopoulos, C.~Bermejo, Z.~Huang, and P.~Hui.
\newblock Mobile augmented reality survey: From where we are to where we go.
\newblock {\em IEEE Access}, 5:6917--6950, 2017.

\bibitem{icra2007Mourikis}
A.I. Mourikis and S.I. Roumeliotis.
\newblock A multi-state constraint {K}alman filter for vision-aided inertial navigation.
\newblock In {\em ICRA}, pages 3565--3572, April 2007.

\bibitem{jfr2010Sibley}
G.~Sibley, L.~Matthies, and G.~Sukhatme.
\newblock Sliding window filter with application to planetary landing.
\newblock {\em JFR}, 27(5):587--608, 2010.

\bibitem{robotics2017Forster}
C.~Forster, Z.~Zhang, M.~Gassner, M.~Werlberger, and D.~Scaramuzza.
\newblock Svo: Semidirect visual odometry for monocular and multicamera systems.
\newblock {\em IEEE Transactions on Robotics}, 33(2):249--265, 2017.

\bibitem{ei2015Valavanis}
Kimon~P Valavanis.
\newblock {Classification of UAVs}.
\newblock In Kimon~P Valavanis and George~J Vachtsevanos, editors, {\em Handbook of Unmanned Aerial Vehicles}. Springer Netherlands, Dordrecht, 2015.

\bibitem{skydio:R1}
{Skydio}.
\newblock \url{https://www.skydio.com/2018/02/introducing-r1/}.

\bibitem{robotics2012Wood}
R~J Wood, B~Finio, M~Karpelson, K~Ma, N~O P{\'e}rez-Arancibia, P~S Sreetharan, H~Tanaka, and J~P Whitney.
\newblock {Progress on {\textquoteleft}pico{\textquoteright} air vehicles}.
\newblock {\em The International Journal of Robotics Research}, 31(11):1292--1302, 2012.

\bibitem{science2013Ma}
J~Bonnet, P~Yin, M~E Ortiz, P~Subsoontorn, and D~Endy.
\newblock {Controlled Flight of a Biologically Inpsired, Insect-Scale Robot}.
\newblock {\em Science}, 340(6132):599--603, 2013.

\bibitem{qualcommCPU}
\url{https://www.qualcomm.com/products/snapdragon/processors/410e}.

\bibitem{nature2015Floreano}
Dario Floreano and Robert~J Wood.
\newblock {Science, technology and the future of small autonomous drones}.
\newblock {\em Nature}, 521(7553):460--466, 2015.

\bibitem{vlsi2013Yoon}
J.~S. Yoon, J.~H. Kim, H.~E. Kim, W.~Y. Lee, S.~H. Kim, K.~Chung, J.~S. Park, and L.~S. Kim.
\newblock A unified graphics and vision processor with a 0.89 /spl mu/w/fps pose estimation engine for augmented reality.
\newblock {\em IEEE Transactions on Very Large Scale Integration (VLSI) Systems}, 21(2):206--216, 2013.

\bibitem{jssc2015hong}
I.~Hong, G.~Kim, Y.~Kim, D.~Kim, B.~G. Nam, and H.~J. Yoo.
\newblock A 27 mw reconfigurable marker-less logarithmic camera pose estimation engine for mobile augmented reality processor.
\newblock {\em IEEE JSSC}, 50(11):2513--2523, 2015.

\bibitem{jssc2018li}
Z.~Li, Q.~Dong, M.~Saligane, B.~Kempke, L.~Gong, Z.~Zhang, R.~Dreslinski, D.~Sylvester, D.~Blaauw, and H.~S. Kim.
\newblock A 1920 $times$ 1080 30-frames/s 2.3 tops/w stereo-depth processor for energy-efficient autonomous navigation of micro aerial vehicles.
\newblock {\em IEEE JSSC}, 53(1):76--90, 2018.

\bibitem{micro2016murray}
S.~Murray, W.~Floyd-Jones, Y.~Qi, G.~Konidaris, and D.~J. Sorin.
\newblock The microarchitecture of a real-time robot motion planning accelerator.
\newblock In {\em IEEE/ACM MICRO}, pages 1--12, 2016.

\bibitem{rss2017Zhang}
Z.~Zhang, A.~Suleiman, L.~Carlone, V.~Sze, and S.~Karaman.
\newblock Visual-inertial odometry on chip: An algorithm-and-hardware co-design approach.
\newblock In {\em RSS}, 2017.

\bibitem{robotics2016Cadena}
C.~Cadena, L.~Carlone, H.~Carrillo, Y.~Latif, D.~Scaramuzza, J.~Neira, I.~Reid, and J.~J. Leonard.
\newblock Past, present, and future of simultaneous localization and mapping: Toward the robust-perception age.
\newblock {\em IEEE Transactions on Robotics}, 32(6):1309--1332, 2016.

\bibitem{rss2015Forster}
C.~Forster, L.~Carlone, F.~Dellaert, and D.~Scaramuzza.
\newblock {IMU} preintegration on manifold for efficient visual-inertial maximum-a-posteriori estimation.
\newblock In {\em RSS}, 2015.

\bibitem{ai1981Lucas}
Bruce~D. Lucas and Takeo Kanade.
\newblock An iterative image registration technique with an application to stereo vision.
\newblock In {\em The International Joint Conference on Artificial Intelligence}, volume~2, pages 674--679, 1981.

\bibitem{cvpr1994Shi}
Jianbo Shi and C.~Tomasi.
\newblock Good features to track.
\newblock In {\em IEEE CVPR}, pages 593--600, 1994.

\bibitem{Hartley2004}
R.~I. Hartley and A.~Zisserman.
\newblock {\em Multiple View Geometry in Computer Vision}.
\newblock Cambridge University Press, ISBN: 0521540518, second edition, 2004.

\bibitem{bmvc2011Kneip}
L.~Kneip, M.~Chli, and R.~Siegwart.
\newblock Robust real-time visual odometry with a single camera and an imu.
\newblock In {\em BMVC}, pages 1--11, 2011.

\bibitem{robots2009Civera}
J.~Civera, O.~G. Grasa, A.~J. Davison, and J.~M.~M. Montiel.
\newblock 1-point ransac for ekf-based structure from motion.
\newblock In {\em IEEE/RSJ International Conference on Intelligent Robots and Systems}, pages 3498--3504, 2009.

\bibitem{it2001Kschischang}
F.~R. Kschischang, B.~J. Frey, and H.~A. Loeliger.
\newblock Factor graphs and the sum-product algorithm.
\newblock {\em IEEE Transactions on Information Theory}, 47(2):498--519, 2001.

\bibitem{robotics2008Kaess}
M.~Kaess, A.~Ranganathan, and F.~Dellaert.
\newblock isam: Incremental smoothing and mapping.
\newblock {\em IEEE Transactions on Robotics}, 24(6):1365--1378, 2008.

\bibitem{delmerico2018benchmark}
Jeffrey Delmerico and Davide Scaramuzza.
\newblock A benchmark comparison of monocular visual-inertial odometry algorithms for flying robots.
\newblock {\em Memory}, 10:20, 2018.

\bibitem{roboticss2016Burri}
Michael Burri, Janosch Nikolic, Pascal Gohl, Thomas Schneider, Joern Rehder, Sammy Omari, Markus~W Achtelik, and Roland Siegwart.
\newblock The euroc micro aerial vehicle datasets.
\newblock {\em The International Journal of Robotics Research}, 2016.

\bibitem{navion_website}
{Navion website}.
\newblock \url{http://navion.mit.edu/}.

\bibitem{robotmag2012Fraundorfer}
F.~Fraundorfer and D.~Scaramuzza.
\newblock Visual odometry: Part ii— matching, robustness, optimization, and applications.
\newblock {\em IEEE Robot. Autom. Mag.}, 19(2):78--90, 2012.






















%\bibitem{AppleARkit}
%\url{https://www.engadget.com/2017/09/12/apple-iphone-x-8-plus-battery-life-pokemon-go/}.

%\bibitem{adi:imu}
%{ADXL345 IMU sensor}.
%\newblock
%  \url{http://www.analog.com/en/products/mems/accelerometers/adxl345.html/}.

%\bibitem{apple:arkit}
%{Apple ARKit}.
%\newblock \url{https://developer.apple.com/arkit/}.

%\bibitem{cv2013Badino}
%H.~Badino, A.~Yamamoto, and T.~Kanade.
%\newblock Visual odometry by multi-frame feature integration.
%\newblock In {\em 2013 IEEE International Conference on Computer Vision Workshops}, pages 222--229, 2013.

%\bibitem{iros2015Bloesch}
%M.~Bloesch, S.~Omari, M.~Hutter, and R.~Siegwart.
%\newblock Robust visual inertial odometry using a direct ekf-based approach.
%\newblock In {\em IEEE/RSJ IROS}, pages 298--304, 2015.

%\bibitem{opencv2000Bradski}
%G.~Bradski.
%\newblock {The OpenCV Library}.
%\newblock {\em Dr. Dobb's Journal of Software Tools}, 2000.

%\bibitem{icra2014Carlone}
%L.~Carlone, Z.~Kira, C.~Beall, V.~Indelman, and F.~Dellaert.
%\newblock Eliminating conditionally independent sets in factor graphs: a unifying perspective based on smart factors.
%\newblock In {\em IEEE International Conference on Robotics and Automation (ICRA)}, 2014.

%\bibitem{isca2016Chen}
%Y.~H. Chen, J.~Emer, and V.~Sze.
%\newblock Eyeriss: A spatial architecture for energy-efficient dataflow for convolutional neural networks.
%\newblock In {\em ACM/IEEE International Symposium on Computer Architecture (ISCA)}, pages 367--379, 2016.

%\bibitem{jssc2017chen}
%Y.~H. Chen, T.~Krishna, J.~S. Emer, and V.~Sze.
%\newblock Eyeriss: An energy-efficient reconfigurable accelerator for deep convolutional neural networks.
%\newblock {\em IEEE JSSC}, 52(1):127--138, 2017.

%\bibitem{aaai2017Clark}
%Ronald Clark, Sen Wang, Hongkai Wen, Andrew Markham, and Niki Trigoni.
%\newblock Vinet: Visual-inertial odometry as a sequence-to-sequence learning problem.
%\newblock In {\em AAAI}, pages 3995--4001, 2017.

%\bibitem{robotics2016Costante}
%G.~Costante, P.~Mancini, M.and~Valigi, and T.~A. Ciarfuglia.
%\newblock Exploring representation learning with cnns for frame-to-frame ego-motion estimation.
%\newblock {\em IEEE Robotics and Automation Letters}, 1(1):18--25, 2016.

%\bibitem{tpami2007Davison}
%A.~J. Davison, I.~D. Reid, N.~D. Molton, and O.~Stasse.
%\newblock Monoslam: Real-time single camera slam.
%\newblock {\em IEEE TPAMI}, 29(6):1052--1067, 2007.

%\bibitem{icra2011Dong}
%T.~C. Dong-Si and A.~I. Mourikis.
%\newblock Motion tracking with fixed-lag smoothing: Algorithm and consistency analysis.
%\newblock In {\em IEEE ICRA}, pages 5655--5662, 2011.

%\bibitem{iccv2013Engel}
%J.~Engel, J.~Sturm, and D.~Cremers.
%\newblock Semi-dense visual odometry for a monocular camera.
%\newblock In {\em IEEE ICCV}, pages 1449--1456, 2013.

%\bibitem{isscc2014Horowitz}
%M.~Horowitz.
%\newblock 1.1 computing's energy problem (and what we can do about it).
%\newblock In {\em IEEE International Solid-State Circuits Conference (ISSCC)}, pages 10--14, 2014.

%\bibitem{ikea}
%{Ikea AR App}.
%\newblock
%  \url{https://www.theverge.com/2017/9/20/16339006/apple-ios-11-arkit-ikea-place-ar-app}.

%\bibitem{robotics2013Indelman}
%V.~Indelman, S.~Wiliams, M.~Kaessand, and F.~Dellaert.
%\newblock “information fusion in navigation systems via factor graph based incremental smoothing.
%\newblock {\em Robot. Auton. Syst.}, 61(8):721--738, 2013.

%\bibitem{robotics2011Jones}
%E.~S. Jones and S.~Soatto.
%\newblock Visual-inertial navigation, mapping and localization: A scalable real-time causal approach.
%\newblock {\em Intl. J. Robot. Res.}, 30(4), 2011.

%\bibitem{cvpr2001Jung}
%S.~H. Jung and C.~J. Taylor.
%\newblock Camera trajectory estimation using inertial sensor measurements and structure from motion results.
%\newblock In {\em IEEE CVPR}, volume~2, pages 732--737, 2001.

%\bibitem{icra2011Kaess}
%M.~Kaess, H.~Johannsson, R.~Roberts, V.~Ila, J.~Leonard, and F.~Dellaert.
%\newblock isam2: Incremental smoothing and mapping with fluid relinearization and incremental variable reordering.
%\newblock In {\em IEEE ICRA}, pages 3281--3288, 2011.

%\bibitem{aiaa2012Keennon}
%Matthew Keennon, Karl Klingebiel, and Henry Won.
%\newblock {Development of the Nano Hummingbird: A Tailless Flapping Wing Micro Air Vehicle}.
%\newblock In {\em AIAA Aerospace Sciences Meeting including the New Horizons Forum and Aerospace Exposition}, Reston, Virigina, June 2012.

%\bibitem{micro2017kim}
%Y.~Kim, D.~Shin, J.~Lee, and H.~J. Yoo.
%\newblock Brain: A low-power deep search engine for autonomous robots.
%\newblock {\em IEEE Micro}, 37(5):11--19, 2017.

%\bibitem{ar2007Klein}
%G.~Klein and D.~Murray.
%\newblock Parallel tracking and mapping for small ar workspaces.
%\newblock In {\em IEEE/ACM International Symposium on Mixed and Augmented Reality}, pages 225--234, 2007.

%\bibitem{icra2014Kneip}
%L.~Kneip and P.~Furgale.
%\newblock Opengv: A unified and generalized approach to real-time calibrated geometric vision.
%\newblock In {\em IEEE International Conference on Robotics and Automation (ICRA)}, 2014.

%\bibitem{cv2015Konda}
%K.~Konda and R.~Memisevic.
%\newblock Learning visual odometry with a convolutional network.
%\newblock {\em International Conference on Computer Vision Theory and Applications}, 2015.

%\bibitem{robotics2015Lenz}
%Lee~H Lenz~I and Saxena A.
%\newblock Deep learning for detecting robotic grasps.
%\newblock {\em The International Journal of Robotics Research}, 34(4--5):705--724, 2015.

%\bibitem{robotics2015Leutenegger}
%S.~Leutenegger, S.~Lynen, M.~Bosse, R.~Siegwart, , and P.~Furgale.
%\newblock Keyframe-based visual-inertial slam using nonlinear optimization.
%\newblock {\em The International Journal of Robotics Research}, 34(3), 2015.

%\bibitem{robotics2016Loianno}
%Giuseppe Loianno, Chris Brunner, Gary McGrath, and Vijay Kumar.
%\newblock Estimation, control, and planning for aggressive flight with a small quadrotor with a single camera and imu.
%\newblock {\em IEEE Robotics and Automation Letters}, 2(2):404--411, 2017.

%\bibitem{tdk:imu}
%{MPU-6050 IMU sensor}.
%\newblock
%  \url{https://www.invensense.com/products/motion-tracking/6-axis/mpu-6050//}.

%\bibitem{robotics2015Artal}
%R.~Mur-Artal, J.~M.~M. Montiel, and J.~D. Tardós.
%\newblock Orb-slam: A versatile and accurate monocular slam system.
%\newblock {\em IEEE Transactions on Robotics}, 31(5):1147--1163, 2015.

%\bibitem{icra2014Nerurkar}
%E.~D. Nerurkar, K.~J. Wu, and S.~I. Roumeliotis.
%\newblock C-klam: Constrained keyframe-based localization and mapping.
%\newblock In {\em IEEE ICRA}, pages 3638--3643, 2014.

%\bibitem{iccv2011Newcombe}
%R.~A. Newcombe, S.~J. Lovegrove, and A.~J. Davison.
%\newblock Dtam: Dense tracking and mapping in real-time.
%\newblock In {\em IEEE ICCV}, pages 2320--2327, 2011.

%\bibitem{icra2014Nikolic}
%J.~Nikolic, J.~Rehder, M.~Burri, P.~Gohl, S.~Leutenegger, P.~T. Furgale, and R.~Siegwart.
%\newblock A synchronized visual-inertial sensor system with fpga pre-processing for accurate real-time slam.
%\newblock In {\em IEEE ICRA}, pages 431--437, 2014.

%\bibitem{oculusBlog}
%{Oculus blog}.
%\newblock \url{https://www.oculus.com/blog/inventing-the-future/}.

%\bibitem{jcv2015Perez}
%Alonso Patron-Perez, Steven Lovegrove, and Gabe Sibley.
%\newblock A spline-based trajectory representation for sensor fusion and rolling shutter cameras.
%\newblock {\em Int. J. Comput. Vision}, 113(3):208--219, 2015.

%\bibitem{ismar2016Rambach}
%J.~R. Rambach, A.~Tewari, A.~Pagani, and D.~Stricker.
%\newblock Learning to fuse: A deep learning approach to visual-inertial camera pose estimation.
%\newblock In {\em IEEE ISMAR}, pages 71--76, 2016.

%\bibitem{robotmag2011Scaramuzza}
%D.~Scaramuzza and F.~Fraundorfer.
%\newblock Visual odometry: Part i—the first 30 years and fundamentals.
%\newblock {\em IEEE Robot. Autom. Mag.}, 18(4):80--92, 2011.

%\bibitem{robotics2004Sterlow}
%D.~Sterlow and S.~Singh.
%\newblock Motion estimation from image and inertial measurements.
%\newblock {\em Intl. J. Robot. Res.}, 23(12):1157--1195, 2004.

%\bibitem{rsj2008Tardif}
%J.~P. Tardif, Y.~Pavlidis, and K.~Daniilidis.
%\newblock Monocular visual odometry in urban environments using an omnidirectional camera.
%\newblock In {\em IEEE/RSJ International Conference on Intelligent Robots and Systems}, pages 2531--2538, 2008.

%\bibitem{report2015Vasilakis}
%E.~Vasilakis.
%\newblock An instruction level energy characterization of arm processors.
%\newblock {\em Technical Report}, 2015.

\end{thebibliography}

\begin{IEEEbiography}[{\includegraphics[width=1in,height=1.25in,clip,keepaspectratio]{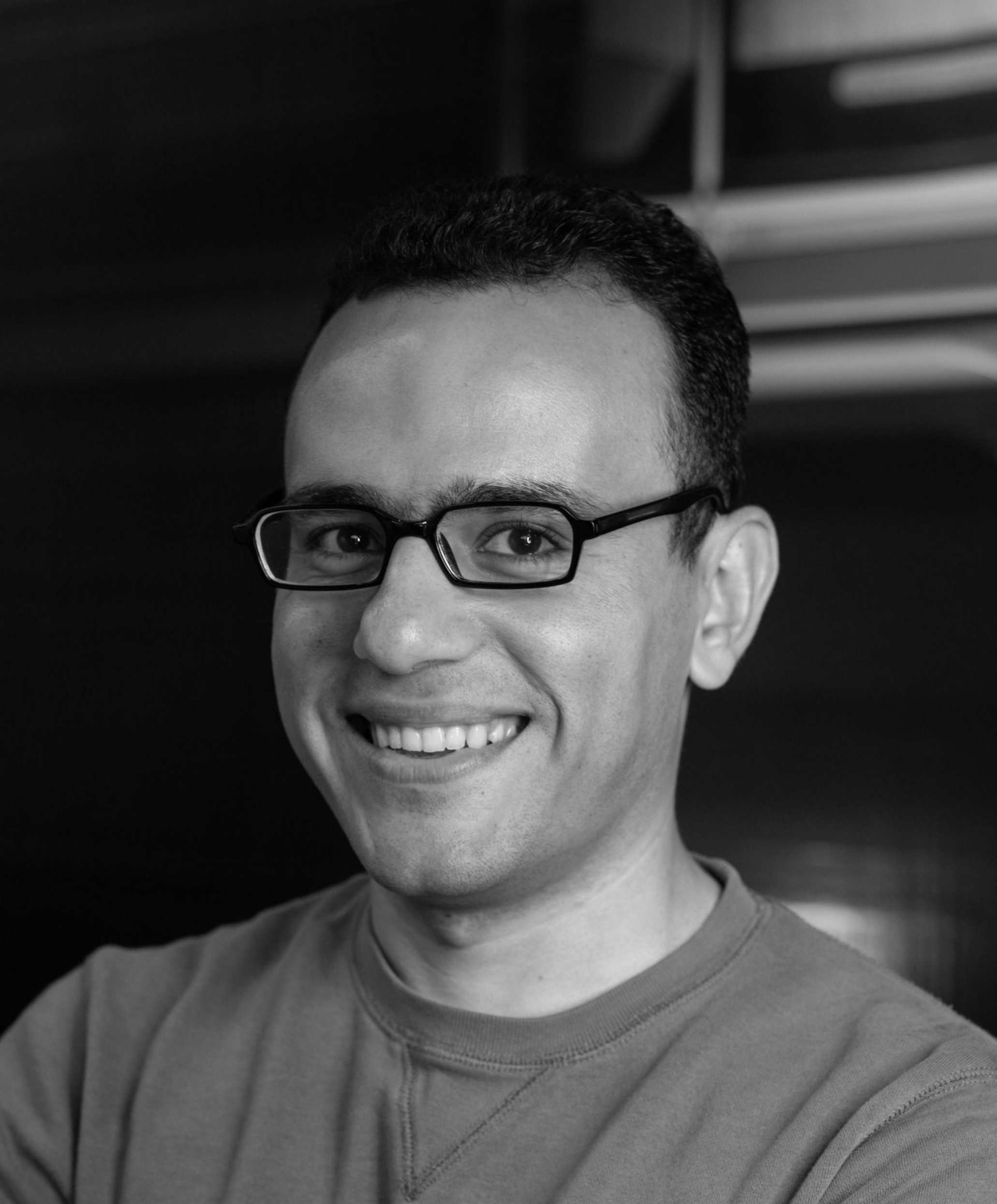}}]{Amr Suleiman}
(S'08--M'18) received the B.S. and M.S. degrees in Electronics and Electrical Communications Engineering from Cairo University, Egypt  in 2008 and 2011 respectively. He received the M.S. and Ph.D. degrees in Electrical Engineering and computer science from MIT in 2013 and 2018 respectively. He is currently working as a research scientist  in Facebook Reality Labs. Amr's research work focuses on developing new energy--efficient implementations for machine vision algorithms (e.g. detection, recognition, and tracking). Amr is a recipient of the Endowed fellowship of the Arab Republic of Egypt, and the 2015 Broadcom Foundation University Research Competition.
\end{IEEEbiography}

\begin{IEEEbiography}[{\includegraphics[width=1in,height=1.25in,clip,keepaspectratio]{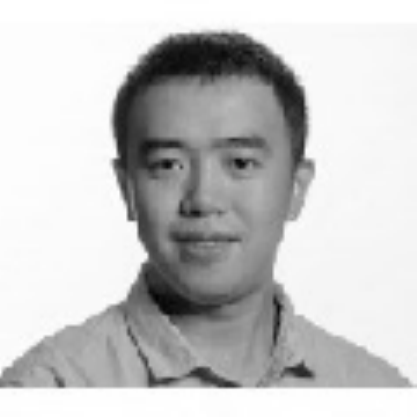}}]{Zhengdong Zhang}
(S'15) received the B.S. in Computer Science in 2011 from Tsinghua University, Beijing, China. He received the M.S. degree in Computer Science from Massachusetts Institute of Technology, Cambridge in 2014. Between 2011 and 2012 he worked in Microsoft Research Asia, Beijing, China, as an Assistant Researcher. He is pursuing the Ph.D. degree under the supervision of Prof. Vivienne Sze. His research interest spans the area of sparsity, low-rank matrix recovery, symmetry/regularity of textures, 3D computer vision, computational photography and vision systems. His current research focuses on the design of energy-efficient vision systems.
\end{IEEEbiography}

\begin{IEEEbiography}[{\includegraphics[width=1in,height=1.25in,clip,keepaspectratio]{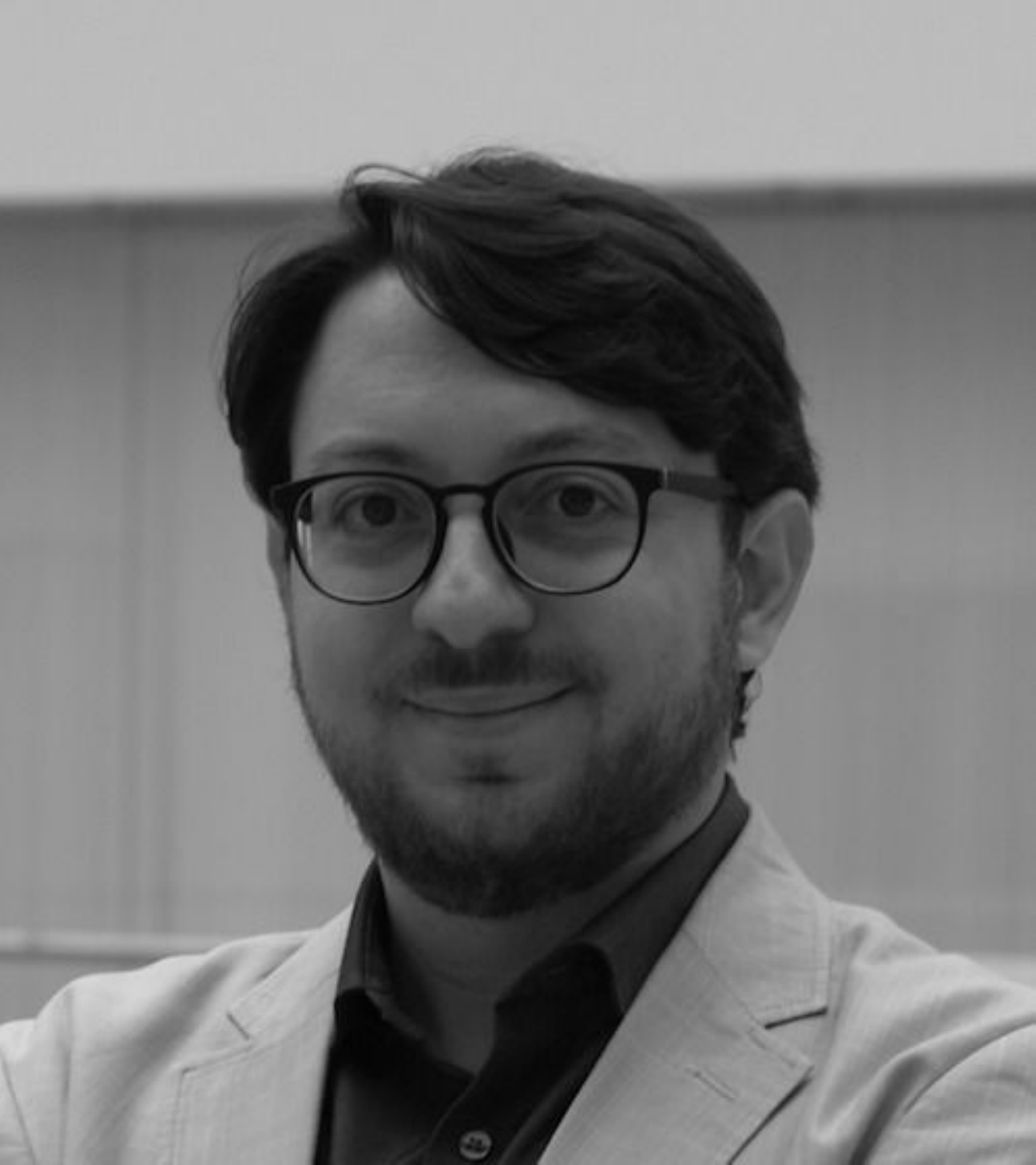}}]{Luca Carlone} is the Charles Stark Draper Assistant Professor in the Department of Aeronautics and Astronautics at the Massachusetts Institute of Technology, and a Principal Investigator in the Laboratory for Information \& Decision Systems (LIDS).  He received his PhD from the Polytechnic University of Turin in 2012. He joined LIDS as a postdoctoral associate (2015) and later as a Research Scientist (2016), after spending two years as a postdoctoral fellow at the Georgia Institute of Technology (2013-2015). His research interests include nonlinear estimation, numerical and distributed optimization, and probabilistic inference, applied to sensing, perception, and decision-making in single and multi-robot systems. His work includes seminal results on certifiably-correct algorithms for localization and mapping, as well as approaches for visual-inertial navigation and distributed mapping. He is the recipient of the 2017 Transactions on Robotics King-Sun Fu Memorial Best Paper Award, and the best paper award at WAFR 2016.
\end{IEEEbiography}

\begin{IEEEbiography}[{\includegraphics[width=1in,height=1.25in,clip,keepaspectratio]{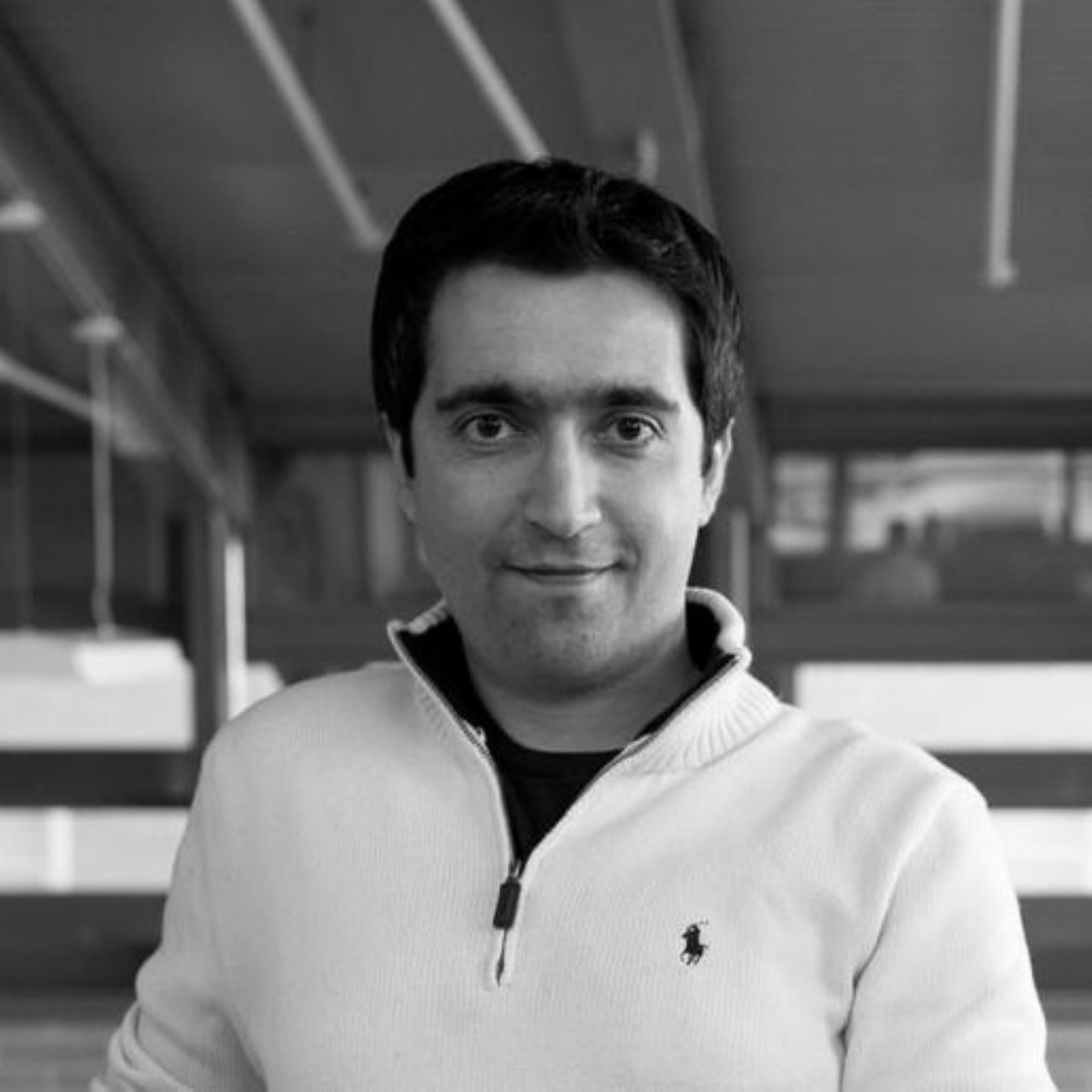}}]{Sertac Karaman} is an Associate Professor of Aeronautics and Astronautics at the Massachusetts Institute of Technology (since Fall 2012). He has obtained B.S. degrees in mechanical engineering and in computer engineering from the Istanbul Technical University, Turkey, in 2007; an S.M. degree in mechanical engineering from MIT in 2009; and a Ph.D. degree in electrical engineering and computer science also from MIT in 2012. His research interests lie in the broad areas of robotics and control theory. In particular, he studies the applications of probability theory, stochastic processes, stochastic geometry, formal methods, and optimization for the design and analysis of high-performance cyber-physical systems. The application areas of his research include driverless cars, unmanned aerial vehicles, distributed aerial surveillance systems, air traffic control, certification and verification of control systems software, and many others. He delivered the the Robotics: Science and Systems Early Career Spotlight Talk in 2017. He is the recipient of an IEEE Robotics and Automation Society Early Career Award in 2017, an Office of Naval Research Young Investigator Award in 2017, Army Research Office Young Investigator Award in 2015, National Science Foundation Faculty Career Development (CAREER) Award in 2014, AIAA Wright Brothers Graduate Award in 2012, and an NVIDIA Fellowship in 2011. He serves as a board member for the Robotics: Science and Systems (RSS) Foundation, as the technical area chair for the robotics area for the IEEE Transactions on Aerospace Electronic Systems, and a co-chair of the IEEE Robotics and Automation Society Technical Committee of Algorithms for the Planning and Control of Robot Motion. 

\end{IEEEbiography}

\begin{IEEEbiography}[{\includegraphics[width=1in,height=1.25in,clip,keepaspectratio]{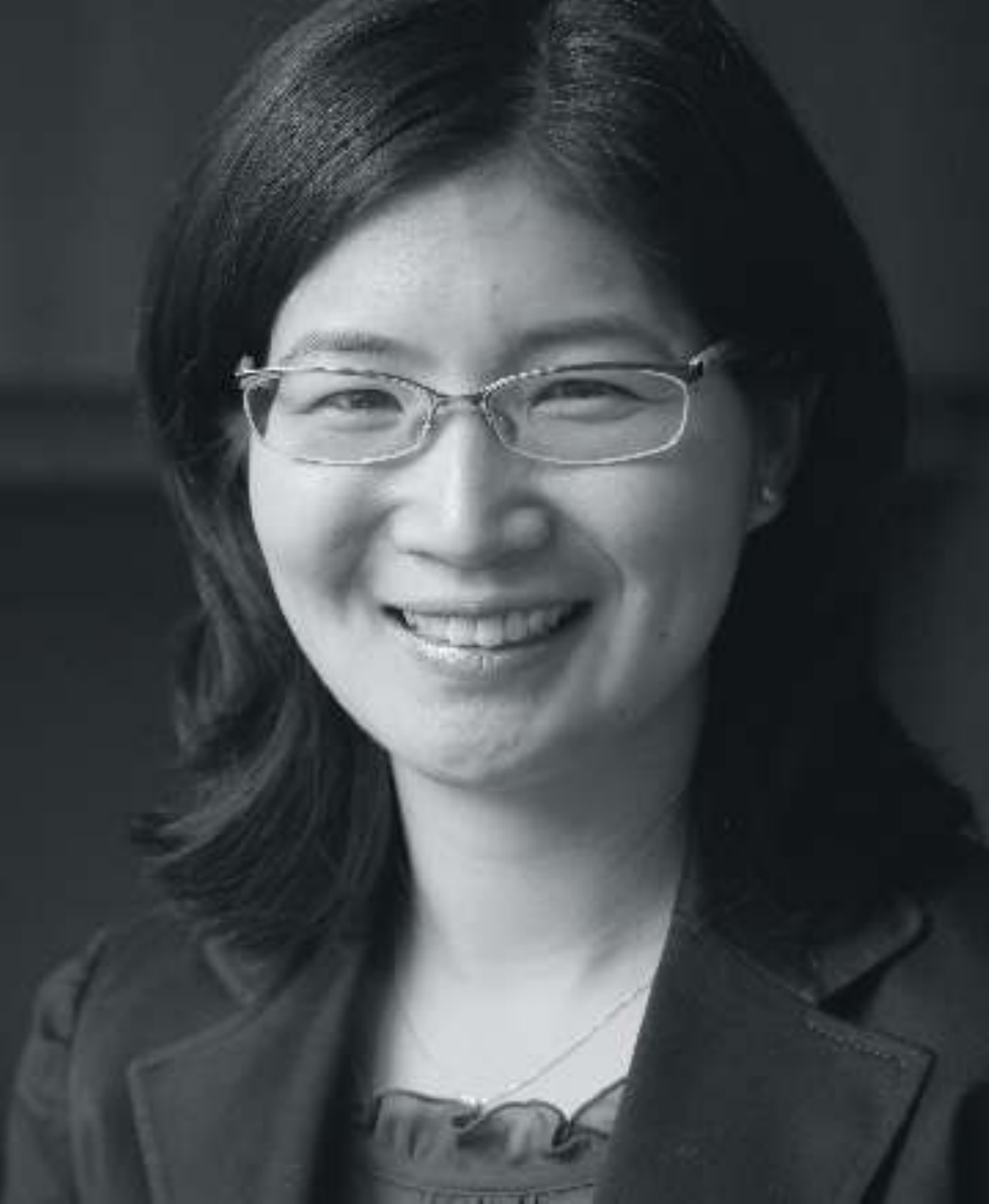}}]{Vivienne Sze}
(S'04--M'10--SM'16) received the B.A.Sc. (Hons) degree in electrical engineering from the University of Toronto, Toronto, ON, Canada, in 2004, and the S.M. and Ph.D. degree in electrical engineering from the Massachusetts Institute of Technology (MIT), Cambridge, MA, in 2006 and 2010 respectively.  In 2011, she received the Jin-Au Kong Outstanding Doctoral Thesis Prize in Electrical Engineering at MIT.

She is an Associate Professor at MIT in the Electrical Engineering and Computer Science Department.  Her research interests include energy-aware signal processing algorithms, and low-power circuit and system design for portable multimedia applications including computer vision, deep learning, autonomous navigation, image processing and video coding. Prior to joining MIT, she was a Member of Technical Staff in the Systems and Applications R\&D Center at Texas Instruments (TI), Dallas, TX, where she designed low-power algorithms and architectures for video coding.  She also represented TI in the JCT-VC committee of ITU-T and ISO/IEC standards body during the development of High Efficiency Video Coding (HEVC), which received a Primetime Emmy Engineering Award.  Within the committee, she was the primary coordinator of the core experiment on coefficient scanning and coding, and has chaired/vice-chaired several ad hoc groups on entropy coding.  She is a co-editor of “High Efficiency Video Coding (HEVC): Algorithms and Architectures” (Springer, 2014).

Prof.\ Sze is a recipient of the 2018 \& 2017 Qualcomm Faculty Award, the 2018 Facebook Faculty Award, the 2016 Google Faculty Research Award, the 2016 AFOSR Young Investigator Research Program Award, the 2016 3M Non-Tenured Faculty Award, the 2014 DARPA Young Faculty Award, the 2007 DAC/ISSCC Student Design Contest Award and a co-recipient of the 2017 CICC Outstanding Invited Paper Award, the 2016 IEEE Micro Top Picks Award and the 2008 A-SSCC Outstanding Design Award.  Prof.\ Sze is a Distinguished Lecturer of the IEEE Solid-State Circuits Society (SSCS), and currently serves on SSCS AdCom and the technical program committees for VLSI Symposium, SysML and MICRO. 
\end{IEEEbiography}

% that's all folks
\end{document}